\documentclass[sigconf,nonacm]{acmart}
\usepackage{subcaption}  
\usepackage{listings}
\setcopyright{none}

\makeatletter
\def\@copyrightyear{}
\def\@copyrightowner{}
\def\@acmISBN{}
\def\@acmDOI{}
\long\def\@copyrightpermission{}

\long\def\footnotetextcopyrightpermission#1{}
\makeatother

\usepackage{titlesec}
\titlespacing*{\section}{0pt}{0.3\baselineskip}{0.2\baselineskip}
\titlespacing*{\subsection}{0pt}{0.2\baselineskip}{0.15\baselineskip}

\usepackage{enumitem}
\setlist{noitemsep,topsep=2pt,parsep=0pt,partopsep=0pt}

\AtBeginDocument{%
  }

\graphicspath{{figures/}{data/figures/}}
\usepackage{booktabs}

\begin{document}

\clearpage

\title{FRAGMENTA: Efficient End-to-end Fragmentation-based Generative Model with Agentic Tuning for Drug Lead Optimization in Small Data Regime}

\author{Yuto Suzuki}
\affiliation{%
  \institution{Department of Computer Science and Engineering\\University of Colorado Denver}
  \city{Denver}
  \state{CO}
  \country{USA}
}
\email{yuto.suzuki@ucdenver.edu}

\author{Paul Awolade}
\affiliation{%
  \institution{Skaggs School of Pharmacy\\University of Colorado Anschutz Medical Campus}
  \city{Aurora}
  \state{CO}
  \country{USA}
}
\email{paul.awolade@cuanschutz.edu}

\author{Daniel V. LaBarbera}
\affiliation{%
  \institution{Skaggs School of Pharmacy\\University of Colorado Anschutz Medical Campus}
  \city{Aurora}
  \state{CO}
  \country{USA}
}
\email{daniel.labarbera@cuanschutz.edu}

\author{Farnoush Banaei-Kashani}
\affiliation{%
  \institution{Department of Computer Science and Engineering\\University of Colorado Denver}
  \city{Denver}
  \state{CO}
  \country{USA}
}
\email{farnoush.banaei-kashani@ucdenver.edu}


\begin{abstract}

  Molecule generation using generative AI models on extremely limited training data is a challenging but crucial problem in drug discovery. Fragment-based models work better with such limited data than atom-based models, but existing fragment-based methods optimize fragment selection independently of the downstream generative process, often overlooking fragments that are most useful for molecule generation. A separate but crucial challenge is that domain expert feedback is even more important with limited data to compensate for this scarcity; however, incorporating such feedback into generative models (via objective functions) requires AI engineering expertise, which slows iteration and sometimes leads to misunderstanding. To address these challenges, we introduce FRAGMENTA, an efficient end-to-end generative framework for drug lead optimization in the small-data regime with two key components: (1) a novel efficient fragment-based molecule generation model that jointly optimizes fragmentation and molecule generation, achieving competitive performance with state-of-the-art (SOTA) methods while training much faster through a simple tabular reward-update mechanism, and (2) an agentic AI system that can automatically refine the objectives of the generative model based on conversational feedback from domain experts. In molecule generation experiments on three small-data datasets (11--104 molecules), LVSEF outperforms SOTA methods in extremely small-data settings and matches SOTA at larger scales, while training ${\sim}16\times$ faster through its tabular reward-update mechanism rather than neural network training. In controlled drug lead optimization experiments across three public protein targets, iterative closed-loop optimization with high-quality feedback improves final-round discovery yield by up to ${\sim}16\%$ relative over one-shot LVSEF-only on kinase; gains are target-dependent and largest when distilled feedback matches the chemistry of the target. Among feedback modalities, Surrogate-Expert--Agent provides competitive performance with the interpretability and knowledge-accumulation advantages of natural language feedback. When deployed in a real-world pharmaceutical laboratory for cancer drug discovery, the Human-Agent configuration of FRAGMENTA identified nearly twice as many molecules with favorable docking scores (i.e., docking score $< -6$) as compared to baseline methods.
\end{abstract}

\maketitle
\vspace{-1.5em}
\section{Introduction}

AI-driven molecule generation has accelerated drug discovery---for example, GENTRL \cite{zhavoronkov2019deep} identified potent candidates in 21 days. Lead optimization, however, typically operates in a small-data regime with fewer than 100--200 validated actives for a given target class \cite{stanley2021fs,guo2022data}. Under these constraints, deep models trained on synthetic corpora often fail to generalize \cite{ma2020pi1m}, and atom-based generators struggle to form complex substructures such as rings. Fragment-based models \cite{jin2020hierarchical,geng2023novo} partially address this by composing reusable atom groups, but their success still depends on which fragments are selected as building blocks.

\begin{figure*}[h]
\begin{center}
\includegraphics[scale=0.2]{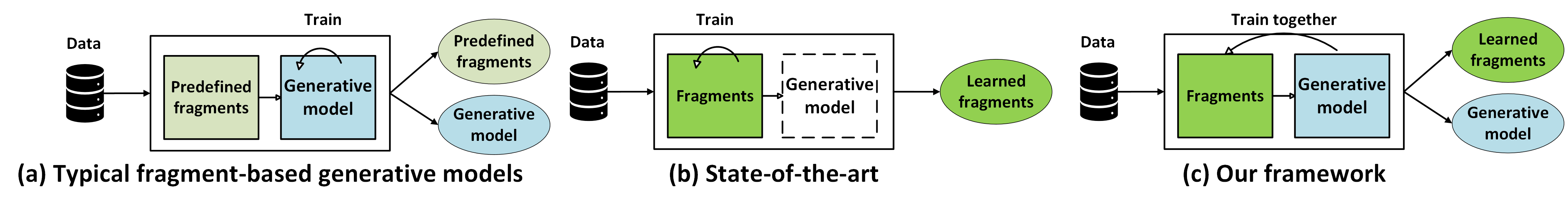}
\end{center}
\caption{Comparison between fragment selection approaches}
\label{concept_diagram}
\end{figure*}

Typical fragment sets are chosen by heuristics such as occurrence frequency, which need not align with generation objectives such as novelty or uniqueness (Figure~\ref{concept_diagram}a). Objective-aware methods such as DEG \cite{guo2022data} improve fragment selection for synthesizability and diversity, yet still ignore the generation process itself (Figure~\ref{concept_diagram}b): fragments are scored under random combination, so a rare motif that is useful only with specific partners can be missed, yielding fragment sets that remain suboptimal for molecule generation.

Aligning generative models with medicinal-chemistry intent is likewise difficult—yet especially important in the small-data regime, where expert feedback must compensate for limited labeled actives. Chemists iteratively critique candidates, but those critiques must be translated by an AI engineer into scoring or reward updates \cite{sundin2022human,choung2023extracting,menke2024metis}. Miscommunication between experts slows design cycles and often loses the intended criteria, so many systems fall back on coarse, one-off signals such as binary preferences rather than iterative adaptation.

In this paper, we introduce \textit{FRAGMENTA}, which couples \textit{LVSEF}---a grammar-aware, end-to-end fragment generative model that jointly selects fragments and connection utilities via a tabular reward-update mechanism (no neural training)---with an agentic controller that clarifies chemist feedback, distills it into executable objectives, and can warm-start fully autonomous Agent-Agent cycles from accumulated knowledge. In the small-data regime (11--200 training molecules), LVSEF outperforms or matches strong fragment baselines while training ${\sim}16\times$ faster than DEG; closed-loop optimization with high-quality feedback raises final-round hit rate by up to ${\sim}16\%$ relative over one-shot LVSEF-only on kinase, with smaller or null gains when feedback is poorly matched to the target; and a seven-month cancer-lab deployment nearly doubles molecules with favorable docking scores (Vina $< -6$) versus human-only tuning. We summarize our contributions as follows:

\begin{itemize}
\item We propose \textit{FRAGMENTA}: LVSEF casts fragment selection as \textit{vocabulary selection} and jointly optimizes fragments with generation, while a multi-agent system clarifies chemist feedback, distills knowledge, and updates objectives in a closed loop.

\item On public small-data benchmarks, LVSEF is competitive while training ${\sim}16\times$ faster than DEG, and closed-loop optimization with high-quality feedback improves final-round hit rate by up to ${\sim}16\%$ relative on kinase.

\item We demonstrate the effectiveness of FRAGMENTA in a real-world pharmaceutical deployment targeting cancer drug discovery. 
\end{itemize}
The LVSEF codebase and agentic system are available at \url{https://tinyurl.com/2a2vecux} and \url{https://tinyurl.com/5365cac4}.

\section{Related Work}

\begin{figure*}[t]  
\vspace{-5pt}
\centering
\begin{tabular}{@{}ccc@{}}
\begin{subfigure}{0.3\textwidth}
    \centering
    \includegraphics[width=\textwidth]{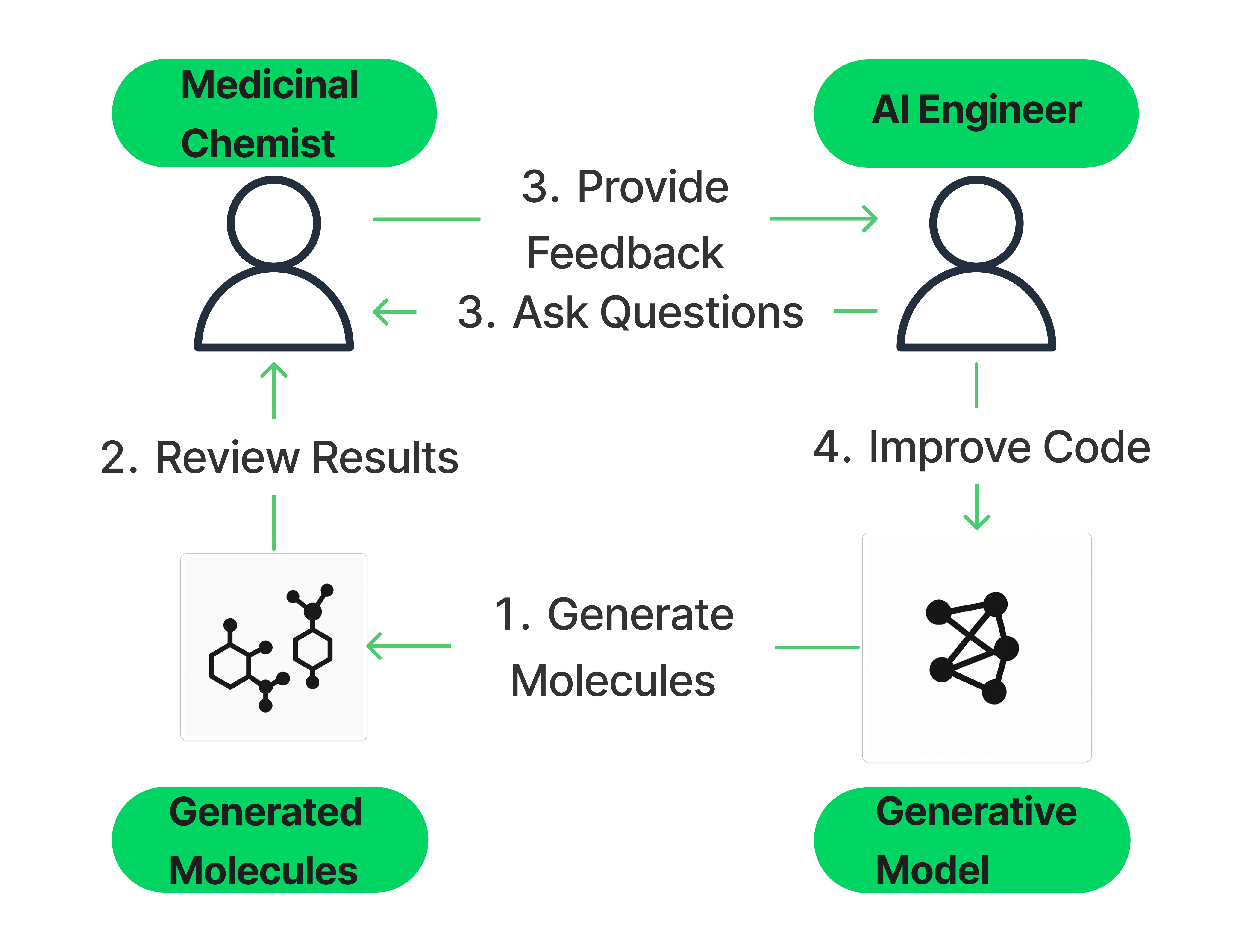}
    \caption{Human-Human}
    \label{human-human}
\end{subfigure} &
\begin{subfigure}{0.3\textwidth}
    \centering
    \includegraphics[width=\textwidth]{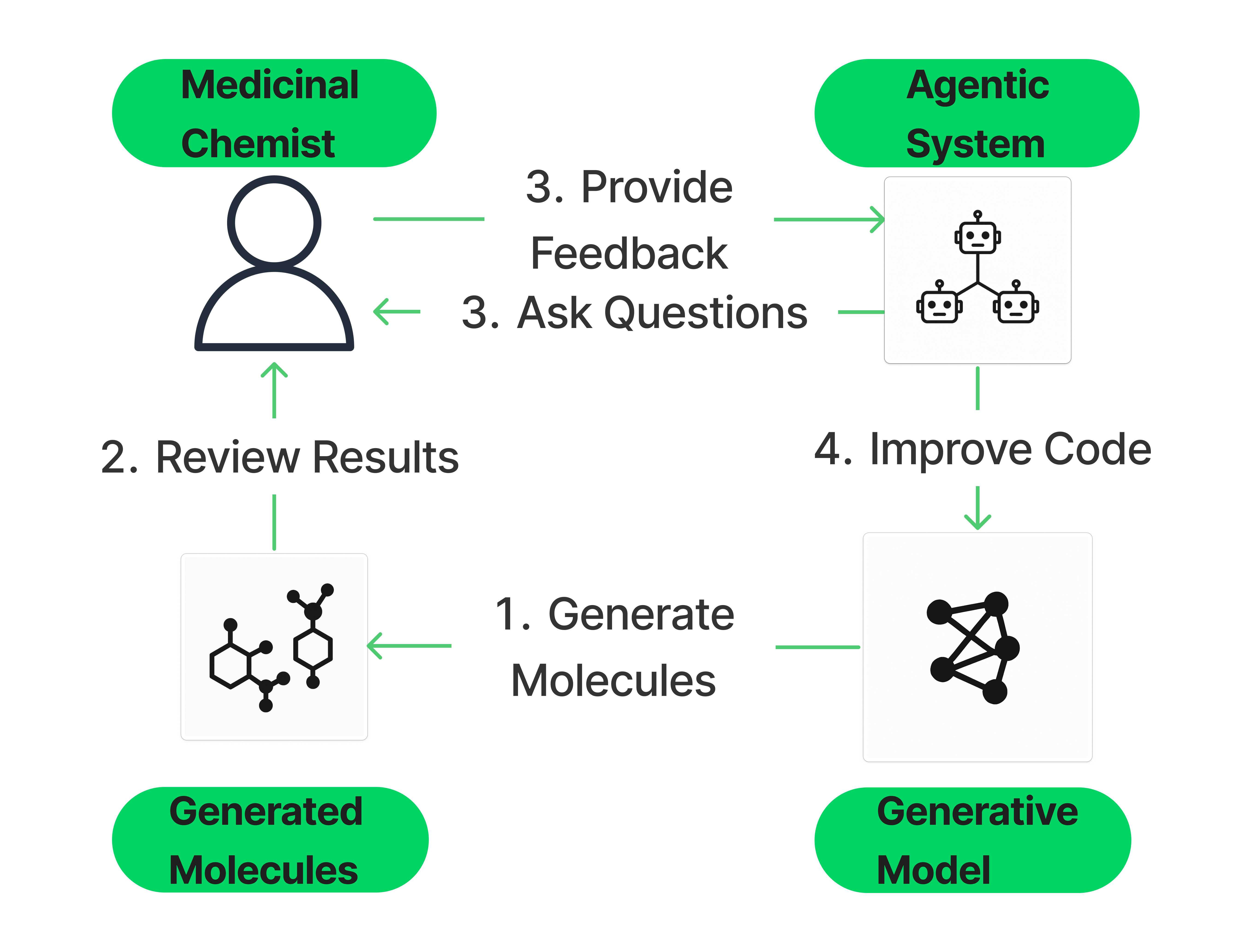}
    \caption{Human-Agent}
    \label{human-agent}
\end{subfigure} &
\begin{subfigure}{0.3\textwidth}
    \centering
    \includegraphics[width=\textwidth]{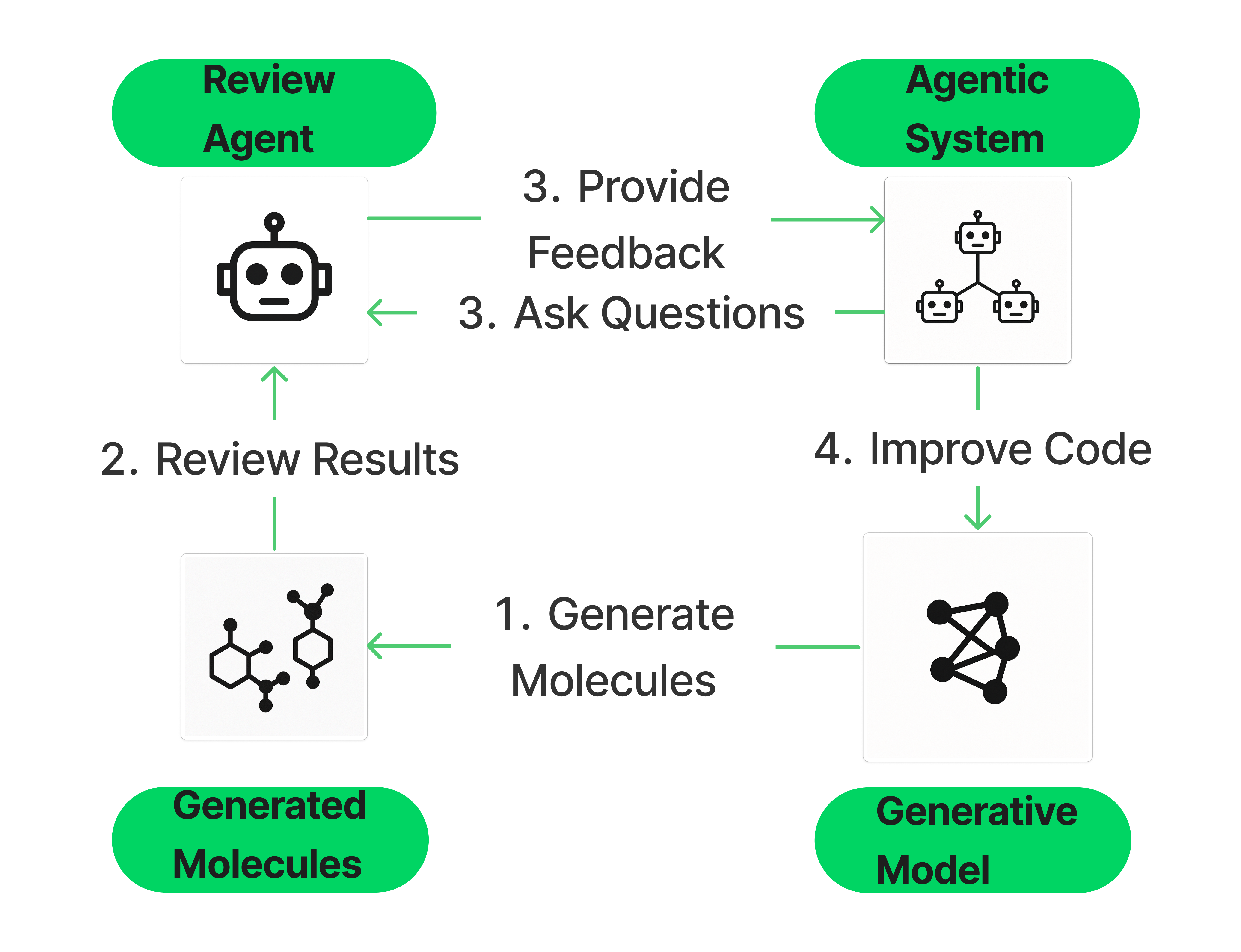}
    \caption{Agent-Agent}
    \label{agent-agent}
\end{subfigure}
\end{tabular}
\caption{Overview of FRAGMENTA configurations. (a) Traditional human-in-the-loop approach with human medicinal chemists and AI engineers. (b) Semi-autonomous system where our agentic framework replaces the AI engineer. (c) Fully autonomous agent-to-agent system where both roles are automated.}
\label{fragmenta_overview}
\vspace{-8pt}
\end{figure*}

\subsection{Generative Models for Molecule Generation}
Molecule generators differ by representation---1D SMILES \cite{smiles}, 2D graphs, or 3D coordinates---and by granularity (atom- vs.\ fragment-based). Sequence models using RNNs/VAEs on SMILES \cite{chenthamarakshan2020cogmol,olivecrona2017molecular,yang2017chemts} often lose structural detail; graph VAEs/GNNs \cite{simonovsky2018graphvae,st2019message}, 3D diffusion \cite{xu2022geodiff,peng2023moldiff}, and related directions (physics-informed \cite{wu2022diffusion}, tree decoding \cite{gao2021amortized}, RL \cite{simm2020symmetry}, flows \cite{kohler2023rigid}, pocket conditioning \cite{liu2022generating}, multi-modal \cite{noh2022path}) recover more geometry. In small-data settings, fragment-based models are often preferred because they reuse chemically meaningful building blocks \cite{jin2018junction,jin2020hierarchical,maziarz2021learning,geng2023novo}, but performance hinges on the fragment vocabulary.

Rule-based vocabularies such as BRICS \cite{degen2008art} and RECAP \cite{lewell1998recap} can be large or biased, while learned decompositions (JT-VAE \cite{jin2018junction}, HierVAE \cite{jin2020hierarchical}, MoLeR \cite{maziarz2021learning}, MiCaM \cite{geng2023novo}) are more flexible yet typically ignore downstream generation goals (e.g., synthesizability, novelty). Task-aware selection \cite{jin2020multi,lee2023drug} still scores fragments in isolation; f-RAG \cite{lee2024molecule} retrieves new motifs without ranking them by generation quality; and DEG \cite{guo2022data} optimizes decomposition with generative objectives but under random fragment combination---missing rare fragments that help only with specific partners. In contrast, we jointly update decomposition and composition from the same downstream rewards, so fragment combinations used in generation are also used for selection (Figure~\ref{concept_diagram}c).

\subsection{Expert-Guided Tuning of Generative AI Models}

Human feedback is widely used to tune generative models. In NLP, RLHF \cite{christiano2017deep} aligns LLMs with preferences (InstructGPT \cite{ouyang2022training}; DPO \cite{rafailov2023direct}), but targets broad preference modeling under heavy human labeling rather than specific expert intent. In molecular design, chemists can interactively reshape scoring functions \cite{sundin2022human}, provide pairwise preferences (MolSkill \cite{choung2023extracting}, Metis \cite{menke2024metis}), or work with tool-augmented LLM workflows (ChemCrow \cite{bran2023chemcrow}); feedback is often coarse (e.g., binary), and humans remain in every step \cite{loeffler2024reinvent}. Conversational systems such as DrugAssist \cite{ye2025drugassist}, ChatDrug \cite{liu2024conversational}, and AgentDrug \cite{le2025agentdrug} edit one molecule at a time instead of retuning the generative objective. Fully agentic loops without chemists (LIDDIA \cite{averly2025liddia}, POLO \cite{wang2025polo}) omit the feedback translation problem that matters in practice. We instead close the loop at the distribution level: clarify typically vague multi-molecule feedback, update generative objectives, and optionally run Agent-Agent using knowledge distilled from prior human experts.

\begin{figure*}[t] 
    \centering
    \includegraphics[scale=0.14]{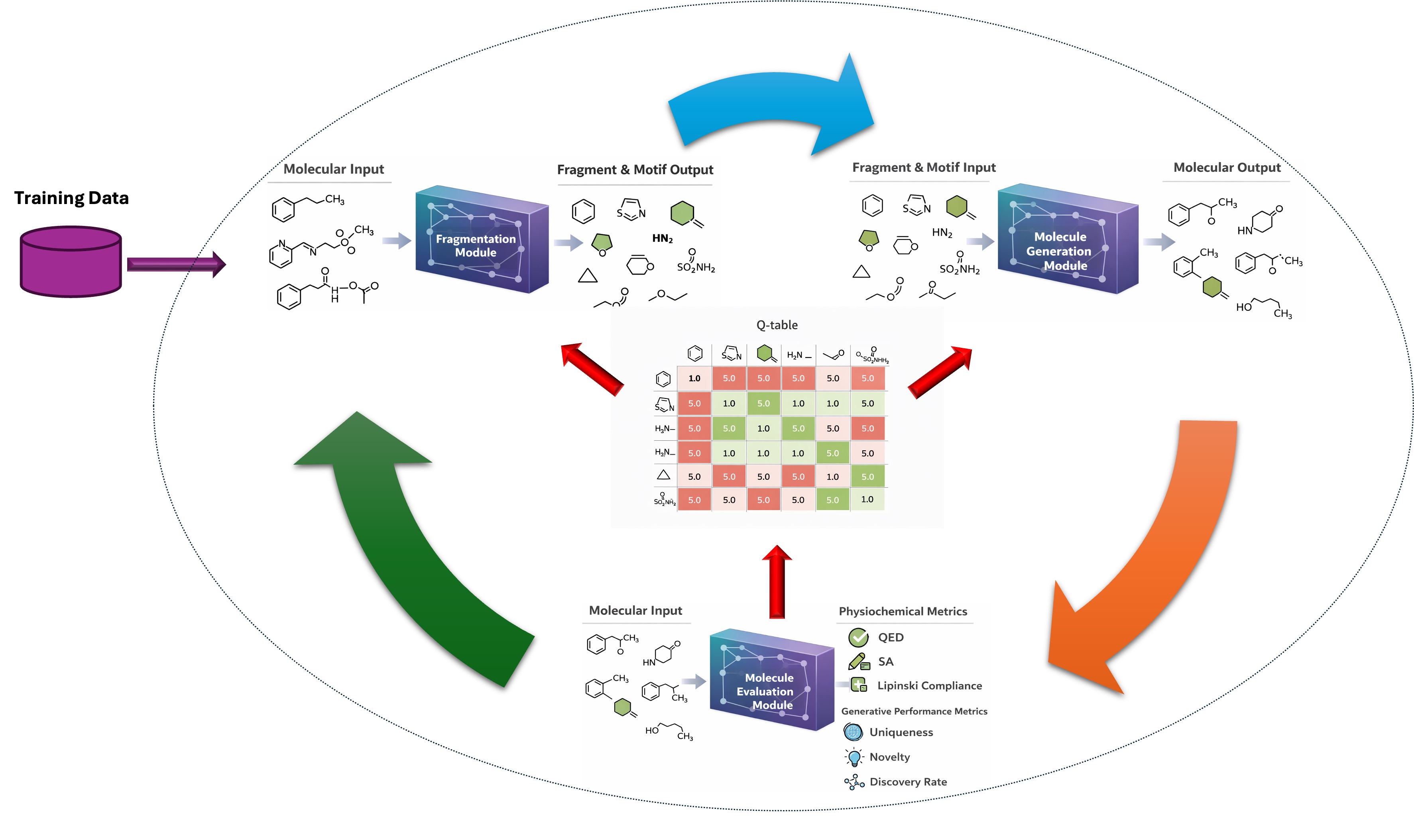}
    \caption{Overview of the LVSEF Framework: (1) Training Data Decomposition: Recursive decomposition of training molecules into high-utility motifs via \textit{Molecular Fragment Ranking} (MFR); (2) Sample Generation: Sequential molecule construction guided by a utility-guided stochastic policy; and (3) Evaluation: Tabular exponential moving-average updates of fragment-connection utilities from batch-level rewards and training-set reconstruction.}
    \label{generative_model_overview}
\end{figure*}

\vspace{-3pt}

\section{FRAGMENTA}

\subsection{Overview}

FRAGMENTA consists of LVSEF, our fragment-based generative model (Section~3.2), and an agentic tuning system (Section~3.3) that iteratively generates candidates and updates the generative objective from expert feedback until quality thresholds (e.g., docking) are met. Tuning can be deployed in three modes (Figure~\ref{fragmenta_overview}):
\begin{itemize}
    \item \underline{Human-Human}: chemists review candidates and AI engineers manually edit objective functions (Figure~\ref{fragmenta_overview}a).
    \item \underline{Human-Agent}: chemists provide feedback to the agentic system, which clarifies intent and updates objectives automatically (Figure~\ref{fragmenta_overview}b).
    \item \underline{Agent-Agent}: a medicinal-chemist agent supplies feedback from distilled knowledge, yielding a fully automated closed loop (Figure~\ref{fragmenta_overview}c).
\end{itemize}






\subsection{Fragment-Based Generative Model (LVSEF)}

\subsubsection{Grammar-aware Vocabulary Selection Problem}

We reformulate fragment selection as a \textit{grammar-aware vocabulary selection} problem, inspired by natural language generation. Just as words are selected based on their expressiveness within a grammar, molecular fragments should be selected to best support the generative process under user-defined objectives (e.g., synthesizability, diversity).

We formalize this problem as follows. Let $S$ be the set of fragments and $P(x \mid S)$ the generative model. The goal is to jointly optimize both $S$ and $P$ to maximize performance across multiple objectives $M_i$ weighted by $\lambda_i$:
\vspace{-5pt}
\[
\max_{S, P} \sum_i \lambda_i M_i(P(x \mid S))
\]
\vspace{-2pt}
We use ``joint optimization'' to mean that fragment decomposition and the composition policy are repeatedly updated from the same downstream molecular rewards, rather than optimizing both through gradient-based end-to-end learning. To make this tractable, we maintain a fragment-connection utility table $\mathcal{U}$, where $U(f_i, f_j)$ stores the learned utility of connecting fragment $f_j$ to fragment $f_i$, and update these utilities with a tabular exponential moving-average rule. This allows us to score and rank fragments based on their utility in downstream molecule generation.

\subsubsection{Model Overview}



Our generative model, LVSEF (short for Learning Vocabulary Selection for Expressive Fragmentation-based molecule generation), implements the procedure above (Figure~\ref{generative_model_overview}). We use graph inputs in our experiments (SMILES also supported). Training proceeds in three steps.

\paragraph{\underline{Step 1: Train Data Decomposition}}
We convert training molecules into a fragment vocabulary via Molecular Fragment Ranking (MFR) and recursive ranking-based cuts, rather than static rules such as BRICS or RECAP.

\hspace{2pt}\noindent\textit{Molecular Fragment Ranking (MFR)} Using the utility table $\mathcal{U}$ from above, we score fragment $f_i$ by its cumulative connection potential:$$MFR(f_i) = \sum_{j=1}^{N} U(f_i, f_j),$$where $N$ is the number of fragments indexed in $\mathcal{U}$. Higher MFR means more high-utility partners; we use MFR to choose cuts, not as a sampling distribution (sampling is Step~2).
 
\hspace{2pt}\noindent\textit{Recursive Ranking-Based Decomposition} At each recursion depth we consider $k$ candidate edge cuts $\mathcal{D}=\{d_1,\ldots,d_k\}$. For cut $d$, let $F_d$ be the resulting fragments and $f_{min,d}=\arg\min_{f\in F_d}\mathrm{size}(f)$ the smaller motif (heavy-atom count). We take$$d^* = \arg\max_{d \in \mathcal{D}} \bigl( MFR(f_{min, d}) \bigr).$$This Max-Min rule prefers high-utility peel-offs and recurses until an atomic-size threshold or no valid cuts remain.

\paragraph{\underline{Step 2: Sample Generation}}
The generative phase of the model treats molecular construction as a sequential decision-making process. Starting from an initial seed fragment, the model iteratively selects subsequent fragments $f_j$ to connect to the current scaffold $f_i$. This selection is governed by a utility-guided stochastic policy. Utilities are initialized to a small exploration constant $\epsilon > 0$ and clipped to remain nonnegative after each update. For valid candidates $f_l \in \mathcal{A}(f_i)$, we define nonnegative sampling weights $w_l = U(f_i, f_l)$ and sample according to$$P(f_j \mid f_i) = \frac{w_j}{\sum_{l \in \mathcal{A}(f_i)} w_l},$$which is a normalized weighting over utilities rather than a softmax. In implementation, generation applies the same principle at the grammar-rule level, combining nonnegative utility values with an exploration schedule over termination rules. The generation process continues recursively until the current fragment contains no available connection points, or a termination condition is met, resulting in the final molecular graph.

\paragraph{\underline{Step 3: Evaluation}}
For a generated batch $\mathcal{B}$, each molecule $M$ receives a composite score $S(M)$ (e.g., QED, SA, binding). Connection $(f_i,f_j)$ aggregates rewards over molecules that contain that edge:$$R_{\text{total}}(f_i, f_j) = \sum_{M \in \mathcal{B}} \mathbb{I}((f_i, f_j) \in E_M) \cdot S(M),$$with $E_M$ the edge set of $M$. Batch aggregation reinforces connections that recur in strong molecules and dampens one-off luck.

\textit{Fragment-Connection Utility Update.} We add a reconstruction bonus $R_{\text{rec}}$ for edges present in the training set $\mathcal{T}$ and update utilities by exponential moving average with learning rate $\alpha\in[0,1]$:$$U_{t+1}(f_i, f_j) = (1-\alpha) U_t(f_i, f_j) + \alpha \left[ R_{\text{total}}(f_i, f_j) + \mathbb{I}((f_i, f_j) \in \mathcal{T}) \cdot R_{\text{rec}} \right].$$This is tabular EMA of observed rewards---not Bellman/Q-learning---and avoids neural training by storing utilities only for observed fragment pairs, which is efficient in the small-data regime.

\subsection{Agentic AI System for Generative Model Tuning}
\begin{figure}[h]
\begin{center}
\includegraphics[scale=0.24]{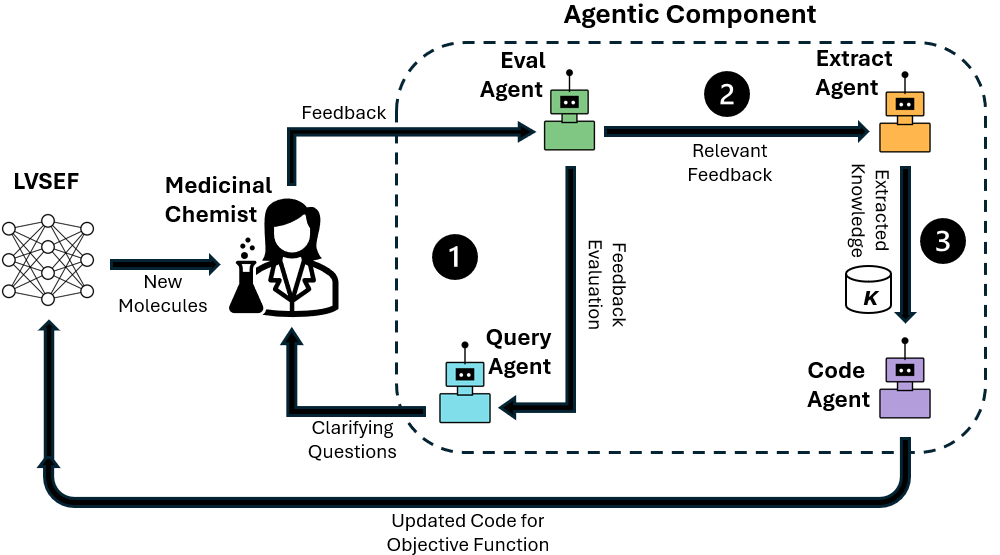}
\end{center}
\caption{Overview of the workflow in agentic system: (1) EvalAgent assesses chemist feedback for novelty and clarity; if insufficient, QueryAgent poses clarifying questions; (2) ExtractAgent distills validated feedback to update the structured Knowledge Base $\mathcal{K}$; and finally, (3) CodeAgent transforms $\mathcal{K}$ into an executable objective function for the LVSEF generative model.}
\label{agentic_system_overview}
\end{figure}

A multi-agent LLM controller turns chemist feedback into LVSEF objectives via the three stages in Figure~\ref{agentic_system_overview}.

\subsubsection{Cyclic Evaluation and Querying to Perceive and Validate Human Feedback: EvalAgent and QueryAgent}
The system first processes the human feedback $L_{expert}$ received in natural language through the \textit{EvalAgent} configured for expert chemical reasoning. Unlike generic LLM calls, the EvalAgent follows a Chain-of-Thought (CoT) reasoning process to deconstruct feedback into core intent. The validation state $V$ is defined by a mapping function:$$V = \text{LLM}_{eval}(L_{expert}, \mathcal{K}) \rightarrow \{s_{clarity}, s_{novelty}\} \in [0, 1]^2$$
where Clarity ($s_{clarity}$) is evaluated based on actionability and quantifiability (e.g., a feedback such as ``make it better'' receives a low score, while ``increase logP to 3.0'' is categorized as high-clarity feedback); and to evaluate Novelty ($s_{novelty}$), the new feedback is cross-referenced with the knowledge base $\mathcal{K}$ to ensure information gain and prevent redundant policy updates. While $\min(s_{clarity}, s_{novelty}) < \tau$, the \textit{QueryAgent} is triggered to ensure the human feedback is not redundant. This agent analyzes the conversation history to identify specific information gaps (e.g., target ranges for logP or specific Lipinski violations) and generates targeted queries to ensure the human feedback eventually meets the threshold $\tau$.

\subsubsection{Knowledge Extraction and Update: ExtractAgent}
Once validated, the human feedback is sent to the \textit{ExtractAgent}, which performs knowledge extraction. The central design element of the agent’s domain understanding is its structured knowledge state representation. The knowledge base $\mathcal{K}$ maintained by the ExtractAgent is a JSON-structured persistent state spanning five predefined categories; the formal schema for these categories is detailed in the Appendix.

\noindent Accordingly, after knowledge extraction the ExtractAgent follows a deduplication-first strategy to update the knowledge state in $\mathcal{K}$:$$\mathcal{K}_{t+1} = \text{Consolidate}(\mathcal{K}_t, \text{Extract}(L_{validated}))$$This prevents state bloat and ensures that similar insights are merged into comprehensive, generalizable statements.

\subsubsection{Objective Function Implementation: CodeAgent}
The \textit{CodeAgent} serves as the system implementer, transforming the knowledge in $\mathcal{K}$ into an executable Python function that will replace the objective function in the LVSEF generative model. To maintain system stability, the CodeAgent is restricted to a strict functional interface:$$\text{def objective(molecules: list[str])} \rightarrow \text{list[float]}$$

\noindent The core design principles governing the CodeAgent's output—including its protocols for score normalization, error handling, and automated self-verification—are detailed in Appendix Section 6.5.

\subsubsection{Agent-to-Agent Autonomous Tuning}
Aside from the main workflow depicted in Figure \ref{agentic_system_overview}, we introduce an \textit{Agent-Agent Mode} as a warm-start autonomous continuation of the Human-Agent phase for the same target and seed (Section~3.1, Figure~\ref{fragmenta_overview}c). In this configuration, the human expert (i.e., the medicinal chemist) is replaced by a surrogate \textit{MedicinalChemistAgent} (not shown in the figure). This agent performs Retrieval-Augmented Generation (RAG) over the knowledge base $\mathcal{K}$ to simulate the expert's preferences. The surrogate PolicyAgent's feedback $L_{agent}$ is generated as:
$L_{agent} = \text{LLM}_{surrogate}(\text{SMILES}(\mathcal{B}), \mathcal{K})$. The system then continues closed-loop optimization using objective updates informed by this surrogate feedback and by knowledge accumulated during the preceding expert-guided rounds. We do not claim that target-specific knowledge transfers across protein targets; Agent-Agent is initialized from knowledge accumulated during the corresponding expert-guided phase for the same target. Evaluating cross-target transfer and cold-start behavior remains future work.

\section{Deployment and Experimental Evaluation}

We evaluate FRAGMENTA in two complementary settings. First, to enable controlled and reproducible comparisons across multiple protein targets and random seeds, we evaluate the framework on public benchmark datasets using an LLM-based medicinal-chemist surrogate that implements the same feedback interface (Section~\ref{sec:benchmark}). Second, we report a prospective deployment conducted over seven months with medicinal chemists in an academic drug-discovery laboratory (Section~\ref{sec:deployment}). Quantitative tables appear below; analysis is in Section~\ref{sec:results}. Full protocol details (datasets, baselines, and secondary metrics) are in Appendix~\ref{sec:appendix_experiment_protocol}.

\subsection{Controlled Benchmark Evaluation}
\label{sec:benchmark}

Benchmark studies use frozen public datasets, fixed computational budgets, and five random seeds; we report mean $\pm$ 95\% confidence intervals unless otherwise stated. In the controlled benchmark, real-time feedback from deployment chemists was not available across the required seeds and targets. We therefore instantiate the expert role using a fixed LLM medicinal-chemist surrogate that exposes the same feedback interface as the conceptual Human-Agent mode in Figure~\ref{fragmenta_overview}b. We refer to this controlled condition as \textbf{Surrogate-Expert--Agent} and reserve \textbf{Human-Agent} for the prospective laboratory deployment involving real medicinal chemists (Section~\ref{sec:deployment}). Agent-Agent in the benchmark is a warm-start autonomous continuation of Surrogate-Expert--Agent for the same target and seed.

\subsubsection{Small-Data Molecular Generation}
We evaluate LVSEF on Chain Extenders (11), Acrylates (32), and a proprietary cancer-lead set (104 actives)\cite{guo2022data}, covering the extreme small-data regime through a more typical lead scale. Baselines are GraphNVP \cite{madhawa2019graphnvp}, HierVAE \cite{jin2020hierarchical}, MiCaM \cite{geng2023novo}, and DEG \cite{guo2022data} (Appendix~\ref{sec:appendix_experiment_protocol}). We evaluate two LVSEF variants and, for end-to-end deployment settings, three feedback configurations (Figure~\ref{fragmenta_overview}). Primary reported metrics include discovery rate, uniqueness, and standard drug-likeness scores; full definitions are in the appendix.

\subsubsection{Target-Specific Lead Optimization (DUD-E)}
We evaluate one representative target from each of three protein families (kinase, GPCR, and nuclear receptor); exact identities and training-active counts appear in Appendix~\ref{sec:appendix_experiment_protocol}. Targets were frozen before runs under a fixed eligibility policy (80--200 actives, receptor availability, and docking verification). Following prior small-data work \cite{guo2022data}, each round generates 100 molecules; one-shot methods train once for 10 epochs, while feedback methods run six rounds. We compare DEG and LVSEF-only as one-shot references, scalar and binary feedback as non-LLM controls, and Surrogate-Expert--Agent and Agent-Agent as closed-loop configurations (full baseline definitions in Appendix~\ref{sec:appendix_experiment_protocol}). Our primary metric is \textbf{hit rate}: the fraction of the 100 raw generated molecules with AutoDock Vina score $< -6$ kcal/mol, measuring discovery yield under a fixed generation budget. Secondary docking and developability metrics are defined in the appendix; best-round and paired-delta tables appear in Appendix~\ref{sec:appendix_dude_tables}.

\label{sec:result_tables}

\begin{table*}[t]
\centering
\caption{Comparative Results with Chain Extenders Dataset}
\label{ce}
\footnotesize
\setlength{\tabcolsep}{3.5pt}
\begin{tabular}{lrrrrrrrrr}
    \toprule
    Model & Dis w/ $\uparrow$ & Dis w/o $\uparrow$ & Valid (\%) $\uparrow$ & RS $\uparrow$ & Unique (\%) $\uparrow$ & Novel (\%) $\uparrow$ & Cham. $\uparrow$ & Div. $\uparrow$ & Mem. (\%) $\uparrow$ \\
    \midrule
    GraphNVP       & -               & -                & 0                   & -           & -                    & -                  & -              & -             & -                 \\
    HierVAE (w/ft) & 1.7             & 3.7              & 100                 & \textbf{0.93} & 5                    & 81                 & \underline{0.56} & 0.77          & 44                \\
    HierVAE (w ft) & 6.3             & 6.4              & 100                 & 0.54        & 31                   & \textbf{98}        & 0.48          & \underline{0.87} & \textbf{98}      \\
    MiCaM & 0            & 0.3              & 100               & \underline{0.67}       & 0.7                  & \textbf{98}        & \textbf{0.81}          & 0.46 & 0     \\
    DEG            & 6.0             & 6.1              & 100                 & 0.51        & 31                   & \textbf{98}        & 0.48          & \underline{0.87} & \underline{97}   \\
    LVSEF (ran)    & \textbf{9.7}    & \textbf{12.1}    & 100                 & 0.55        & \textbf{47}          & 96                 & \underline{0.56} & \textbf{0.88} & 87                \\
    LVSEF (bal)    & \underline{8.3} & \underline{8.3}  & 100                 & 0.62 & \underline{39}       & 95                 & 0.53          & 0.86          & 92                \\
    \bottomrule
\end{tabular}
\end{table*}

\begin{table*}[t]
\centering
\caption{Comparative Results with Acrylates Dataset}
\label{acry}
\footnotesize
\setlength{\tabcolsep}{3.5pt}
\begin{tabular}{lrrrrrrrrr}
    \toprule
    Model & Dis w/ $\uparrow$ & Dis w/o $\uparrow$ & Valid (\%) $\uparrow$ & RS $\uparrow$ & Unique (\%) $\uparrow$ & Novel (\%) $\uparrow$ & Cham. $\uparrow$ & Div. $\uparrow$ & Mem. (\%) $\uparrow$ \\
    \midrule
    GraphNVP       & -               & -                & 0                   & -           & -                    & -                  & -              & -             & -                 \\
    HierVAE (w/ft) & 0               & 2.6              & 100                 & \textbf{0.98} & 3                    & \textbf{100}       & 0.44          & 0.67          & 0                 \\
    HierVAE (w.ft) & 0               & 7.8              & 100                 & 0.56 & 32                   & 95                 & 0.20          & 0.77          & 0                 \\
        MiCaM & 0               & 0.8              & 100                 & \underline{0.67} & 1.2                  & \textbf{100}                & \textbf{0.73}          & 0.58         & 0                 \\
    DEG            & 3.9             & 13.6             & 100                 & 0.33        & \underline{70}       & \textbf{100}       & \underline{0.63} & \textbf{0.87} & 45                \\
    LVSEF (ran)    & \underline{6.2} & \underline{19.2} & 100                 & 0.35        & \textbf{72}          & \textbf{100}       & 0.60 & \underline{0.84} & \textbf{51}      \\
    LVSEF (bal)    & \textbf{6.4}    & \textbf{20.3}    & 100                 & 0.44        & 63                   & \textbf{100}       & 0.52          & \underline{0.84} & \textbf{51}      \\
    \bottomrule
\end{tabular}
\end{table*}

\begin{table*}[t]
\centering
\caption{Comparative Results with Our Internal Dataset}
\label{internal_lvsef}
\footnotesize
\setlength{\tabcolsep}{3.5pt}
\begin{tabular}{lrrrrrrrr}
    \toprule
    Model & Valid (\%) $\uparrow$ & Unique (\%) $\uparrow$ & Novel (\%) $\uparrow$ & Diversity $\uparrow$ & QED $\uparrow$ & SA $\downarrow$ & Lipinski (\%) $\uparrow$ & Scaffold Div. $\uparrow$ \\
    \midrule
    GraphNVP        & -                   & -                    & -                   & -                  & -            & -           & -                      & -                                         \\
    HierVAE(w/o ft)         & \textbf{100.0}      & 51.0                 & 92.2                & \textbf{0.864}     & 0.565        & \underline{2.411} & 99.0                   & 0.27                   \\
    MiCaM           & \textbf{100.0}      & 3.0                  & 66.7                & 0.097              & 0.605        & \textbf{2.278} & \textbf{100.0}       & 0.03                                     \\
    DEG       & \textbf{100.0}      & \underline{96.0}     & \textbf{100.0}      & 0.838              & \textbf{0.754} & 2.790      & \textbf{100.0}        & \underline{0.88}              \\
    LVSEF (ran)     & \textbf{100.0}      & \textbf{99.0}        & \textbf{100.0}      & \underline{0.849}  & \underline{0.730} & 2.706    & \textbf{100.0}        & \textbf{0.90}                     \\
    \bottomrule
\end{tabular}
\end{table*}

\begin{table*}[t]
  \caption{Final-round performance on Kinase (FAK1) (round~6 for feedback conditions; round~1 for one-shot baselines). Five seeds ($n{=}5$). Mean $\pm$ 95\% CI half-width.}
  \label{tab:final_round_kinase}
  \centering
  \footnotesize
  \setlength{\tabcolsep}{3.5pt}
  \begin{tabular}{l r r r r r r r r}
    \toprule
    Condition & Hit rate (\%) $\uparrow$ & Hits ($\leq -6$) $\uparrow$ & Top-10 Vina $\downarrow$ & Best Vina $\downarrow$ & QED $\uparrow$ & SA $\downarrow$ & Valid (\%) $\uparrow$ & Unique (\%) $\uparrow$ \\
    \midrule
    Surrogate-Expert--Agent & \textbf{32.2 ± 6.2\%} & \textbf{32.2 ± 6.2} & -8.44 ± 0.38 & \underline{-10.01 ± 0.92} & \textbf{0.541 ± 0.018} & \underline{0.529 ± 0.021} & 100.0 & 99.6 $\pm$ 0.7 \\
    Agent-Agent & 29.2 ± 3.9\% & 29.2 ± 3.9 & \textbf{-8.51 ± 0.33} & -9.88 ± 0.65 & 0.532 ± 0.017 & 0.531 ± 0.021 & 100.0 & \underline{99.8 $\pm$ 0.6} \\
    Binary feedback & \underline{30.0 ± 7.7\%} & \underline{30.0 ± 7.7} & -8.38 ± 0.35 & -9.53 ± 0.85 & 0.535 ± 0.032 & 0.536 ± 0.016 & 100.0 & \textbf{100.0 $\pm$ 0.0} \\
    Scalar feedback & 28.8 ± 7.7\% & 28.8 ± 7.7 & \underline{-8.46 ± 0.44} & \textbf{-10.34 ± 0.96} & 0.521 ± 0.021 & 0.535 ± 0.021 & 100.0 & \textbf{100.0 $\pm$ 0.0} \\
    DEG & 26.6 ± 6.5\% & 26.6 ± 6.5 & -8.13 ± 0.36 & -9.51 ± 0.99 & 0.502 ± 0.019 & \textbf{0.513 ± 0.028} & 100.0 & \textbf{100.0 $\pm$ 0.0} \\
    LVSEF-only & 27.8 ± 7.6\% & 27.8 ± 7.6 & -8.30 ± 0.21 & -9.71 ± 1.35 & \underline{0.537 ± 0.031} & 0.535 ± 0.031 & 100.0 & \textbf{100.0 $\pm$ 0.0} \\
    \bottomrule
  \end{tabular}
\end{table*}

\begin{table*}[t]
  \caption{Final-round performance on GPCR (CXCR4) (round~6 for feedback conditions; round~1 for one-shot baselines). Five seeds ($n{=}5$). Mean $\pm$ 95\% CI half-width.}
  \label{tab:final_round_gpcr}
  \centering
  \footnotesize
  \setlength{\tabcolsep}{3.5pt}
  \begin{tabular}{l r r r r r r r r}
    \toprule
    Condition & Hit rate (\%) $\uparrow$ & Hits ($\leq -6$) $\uparrow$ & Top-10 Vina $\downarrow$ & Best Vina $\downarrow$ & QED $\uparrow$ & SA $\downarrow$ & Valid (\%) $\uparrow$ & Unique (\%) $\uparrow$ \\
    \midrule
    Surrogate-Expert--Agent & \underline{28.4 ± 4.0\%} & \underline{28.4 ± 4.0} & -8.05 ± 0.18 & -9.11 ± 0.14 & \underline{0.502 ± 0.015} & 0.508 ± 0.014 & 100.0 & 98.6 $\pm$ 1.4 \\
    Agent-Agent & \textbf{30.4 ± 5.6\%} & \textbf{30.4 ± 5.6} & \underline{-8.43 ± 0.22} & \textbf{-10.02 ± 1.01} & 0.501 ± 0.032 & 0.525 ± 0.018 & 100.0 & \textbf{99.6 $\pm$ 0.7} \\
    Binary feedback & 25.2 ± 4.3\% & 25.2 ± 4.3 & -7.98 ± 0.09 & -9.11 ± 0.19 & 0.498 ± 0.032 & \underline{0.505 ± 0.021} & 100.0 & \textbf{99.6 $\pm$ 0.7} \\
    Scalar feedback & 27.4 ± 4.4\% & 27.4 ± 4.4 & -8.16 ± 0.47 & -9.58 ± 1.37 & 0.495 ± 0.023 & 0.526 ± 0.024 & 100.0 & \underline{99.2 $\pm$ 2.2} \\
    DEG & 28.2 ± 6.3\% & 28.2 ± 6.3 & \textbf{-8.43 ± 0.41} & \underline{-9.60 ± 0.35} & \textbf{0.507 ± 0.010} & 0.523 ± 0.024 & 100.0 & 99.0 $\pm$ 1.8 \\
    LVSEF-only & 26.4 ± 5.3\% & 26.4 ± 5.3 & -8.15 ± 0.36 & -9.09 ± 0.48 & 0.488 ± 0.011 & \textbf{0.504 ± 0.019} & 100.0 & \textbf{99.6 $\pm$ 0.7} \\
    \bottomrule
  \end{tabular}
\end{table*}

\begin{table*}[t]
  \caption{Final-round performance on Nuclear receptor (MCR) (round~6 for feedback conditions; round~1 for one-shot baselines). Five seeds ($n{=}5$). Mean $\pm$ 95\% CI half-width.}
  \label{tab:final_round_nuclear_receptor}
  \centering
  \footnotesize
  \setlength{\tabcolsep}{3.5pt}
  \begin{tabular}{l r r r r r r r r}
    \toprule
    Condition & Hit rate (\%) $\uparrow$ & Hits ($\leq -6$) $\uparrow$ & Top-10 Vina $\downarrow$ & Best Vina $\downarrow$ & QED $\uparrow$ & SA $\downarrow$ & Valid (\%) $\uparrow$ & Unique (\%) $\uparrow$ \\
    \midrule
    Surrogate-Expert--Agent & 18.4 ± 3.4\% & 18.4 ± 3.4 & -7.66 ± 0.27 & -8.89 ± 0.28 & 0.519 ± 0.017 & 0.527 ± 0.021 & 100.0 & \underline{99.8 $\pm$ 0.6} \\
    Agent-Agent & 18.2 ± 7.5\% & 18.2 ± 7.5 & -7.48 ± 0.93 & -8.47 ± 1.43 & \underline{0.526 ± 0.028} & 0.524 ± 0.024 & 100.0 & \textbf{100.0 $\pm$ 0.0} \\
    Binary feedback & \textbf{21.2 ± 4.2\%} & \textbf{21.2 ± 4.2} & -7.64 ± 0.36 & -8.67 ± 0.84 & 0.520 ± 0.017 & 0.518 ± 0.045 & 100.0 & \underline{99.8 $\pm$ 0.6} \\
    Scalar feedback & \underline{19.6 ± 4.4\%} & \underline{19.6 ± 4.4} & \textbf{-7.84 ± 0.31} & \underline{-9.28 ± 0.69} & \textbf{0.534 ± 0.023} & \underline{0.516 ± 0.020} & 100.0 & \underline{99.8 $\pm$ 0.6} \\
    DEG & 17.4 ± 2.4\% & 17.4 ± 2.4 & -7.42 ± 0.19 & -8.74 ± 0.68 & 0.520 ± 0.007 & 0.520 ± 0.026 & 100.0 & \textbf{100.0 $\pm$ 0.0} \\
    LVSEF-only & 18.0 ± 5.6\% & 18.0 ± 5.6 & \underline{-7.74 ± 0.40} & \textbf{-9.61 ± 1.60} & 0.513 ± 0.023 & \textbf{0.509 ± 0.023} & 100.0 & \underline{99.8 $\pm$ 0.6} \\
    \bottomrule
  \end{tabular}
\end{table*}

\begin{table*}[t]
\centering
\small
\caption{Agent-based Molecular Generation Results}
\label{end_to_end_results}
\begin{tabular}{lrrrrrrrrr}
    \toprule
    \textbf{Model} & \textbf{Valid (\%) ↑} & \textbf{Unique (\%) ↑} & \textbf{Novel (\%) ↑} & \textbf{Diversity ↑} & \textbf{QED ↑} & \textbf{SA ↓} & \textbf{Lipinski (\%) ↑} & \textbf{Scaffold Div. ↑} & \textbf{Dock <-6 ↑} \\
    \midrule
    Baseline     & \textbf{100.0}      & 99.0                 & \textbf{100.0}      & \underline{0.849}              & 0.730        & \underline{2.706} & \textbf{100.0}        & \underline{0.90}                   & -                 \\
    Human-Human     & \textbf{100.0}      & 99.0                 & \textbf{100.0}      & \textbf{0.867}     & 0.738        & 2.897       & \textbf{100.0}        & 0.87                   & 7                 \\
    Human-Agent        & \textbf{100.0}      & \textbf{100.0}       & \textbf{100.0}      & \underline{0.849}              & \textbf{0.785} & 2.723     & \textbf{100.0}        & \textbf{0.93}          & \textbf{13}       \\
    Agent-Agent        & \textbf{100.0}      & \textbf{100.0}       & \textbf{100.0}      & 0.840  & \underline{0.754} & \textbf{2.666} & \textbf{100.0}        & 0.86      & \underline{11}         \\
    \bottomrule
\end{tabular}
\end{table*}

\subsection{Deployment Overview}
\label{sec:deployment}
We deployed FRAGMENTA over seven months in an academic medicinal chemistry laboratory for a cancer lead-optimization campaign against a proprietary protein target (IP details omitted). Two medicinal chemists participated throughout (postdoctoral primary user and professor supervisor), starting from 104 experimentally validated actives. Weekly closed-loop Human-Human and Human-Agent cycles produced candidates for expert review; workflow details (CLI integration, top-100 ranking, six feedback rounds) are in Appendix~\ref{sec:appendix_deployment}.

\section{Results and Analysis}
\label{sec:results}

\subsection{Small-Data Molecular Generation}
\label{sec:results_small_data}

We evaluated LVSEF across publicly available small monomer datasets (Chain Extenders and Acrylates) and a proprietary internal cancer-lead dataset (104 actives).

As shown in Tables~\ref{ce} and~\ref{acry}, LVSEF consistently outperformed DEG in discovery rate on both public datasets. These sets contain fewer than 100 molecules, and LVSEF generated novel, synthesizable molecules belonging to the same monomer class as the training data. LVSEF(ran) outperformed LVSEF(bal), suggesting that random exploration can be more effective when the searchable space is tightly constrained.

The discovery-rate gains over DEG can be explained by the ablations in Appendix~(Figures~\ref{fig:ablation_generation} and~\ref{fig:ablation_mfr}; Table~\ref{num_atoms_fragments}). First, on Acrylates, our utility-guided generation outperforms random combination under the same fragment set, indicating that learned fragment connections help when there is enough combinatorial search space. Second, under matched random combination, LVSEF fragments still outperform DEG, suggesting that the selected fragments are more useful for the discovery objectives; LVSEF fragments also average 4.2 atoms versus 3.0 for DEG on Acrylates, consistent with retaining larger, reusable motifs rather than over-fragmented building blocks.

Across all models, HierVAE and MiCaM achieved relatively high assay scores but low uniqueness and scaffold diversity, consistent with convergence to local optima in extreme low-data regimes (11--104 molecules). On the internal dataset (Table~\ref{internal_lvsef}), LVSEF(ran) achieved the best uniqueness and scaffold diversity while remaining competitive on QED and drug-likeness metrics; DEG reached higher QED but with lower diversity. 

\subsection{DUD-E Target-Specific Lead Optimization}
\label{sec:dude_results}

Final-round results differ substantially across target families (Tables~\ref{tab:final_round_kinase}--\ref{tab:final_round_nuclear_receptor}). Kinase (FAK1) is the most favorable setting: Surrogate-Expert--Agent reaches 32.2\% raw hit rate, ahead of automated feedback and one-shot baselines at 27--30\%. GPCR (CXCR4) is more competitive: Agent-Agent attains the highest final-round hit rate (30.4\%), slightly above Surrogate-Expert--Agent (28.4\%) and DEG (28.2\%). Nuclear receptor (MCR) is the hardest target, with hit rates near 17--21\%; binary feedback leads at 21.2\%, while Surrogate-Expert--Agent (18.4\%) trails the stronger automated baselines on this family. All conditions maintain 100\% validity; differences among methods are often within overlapping confidence intervals at $n{=}5$ seeds per target. Closed-loop feedback improves discovery over one-shot generation: hit rate generally rises across optimization rounds for feedback conditions, whereas one-shot DEG and LVSEF-only remain flat (round-by-round and per-target views in Appendix~\ref{sec:appendix_dude_tables}, Figures~\ref{fig:trajectory_hit} and~\ref{fig:final_by_target}).

Tracing how objectives change across feedback rounds helps explain why Surrogate-Expert--Agent works well on kinase but underperforms on nuclear receptors (Tables~\ref{tab:objective_trajectory_kinase} and~\ref{tab:objective_trajectory_mcr}). On kinase, later rounds add increasingly target-specific structure to the objective---hinge-binding SMARTS rewards, warhead filters, and ADME shaping---that align with docking-relevant chemistry; the higher hit rate (${\sim}25$--33\%) also provides richer per-round signal for the LLM surrogate to learn from, and kinase inhibitor design is well represented in the model's medicinal-chemistry priors. On nuclear receptors, by contrast, hit rates are lower (${\sim}17$--21\%), so each round yields sparser feedback; without strong target-specific guidance, the LLM tends to fall back on generic developability rules that can conflict with the charged chemotypes common among MCR actives. Together, these patterns suggest that our Surrogate-Expert--Agent is effective at operationalizing expert feedback into code, but its benefit depends heavily on whether that feedback---or the surrogate's domain knowledge---is well matched to the target.

The distilled knowledge bases in Appendix~\ref{sec:appendix_knowledge} further clarify when Agent-Agent helps. On GPCR (CXCR4), Surrogate-Expert--Agent accumulates a clear, target-compatible prior---a cationic nitrogen for a salt-bridge plus rigid hydrophobic scaffolds---and Agent-Agent warm-started from that knowledge attains the best final- and best-round hit rates among conditions (Tables~\ref{tab:final_round_gpcr} and~\ref{tab:best_round_gpcr}). On kinase, the LLM surrogate already has strong hinge-binding priors that Surrogate-Expert--Agent encodes directly into objectives, so additional warm-start knowledge gives Agent-Agent little further advantage over Surrogate-Expert--Agent. On nuclear receptor, Surrogate-Expert--Agent itself fails to learn a successful prior (generic charge/developability rules that conflict with charged actives), so Agent-Agent inherits that misaligned knowledge and likewise does not lead. Altogether, Agent-Agent depends on the quality of the knowledge it inherits: when Surrogate-Expert--Agent learns the right chemistry, autonomous continuation can improve upon it; when that knowledge is wrong or redundant with strong base priors, the gain disappears. This also suggests that Agent-Agent should benefit even more when the warm-start knowledge comes from a real medicinal chemist rather than an LLM surrogate.

\subsubsection{From Review Text to Training Objectives}
\label{sec:objective_learning}

A central capability of FRAGMENTA is translating natural-language medicinal-chemistry review into an executable LVSEF objective. On kinase (FAK1, seed~0), the LLM surrogate reviews docking-ranked SMILES each round; the CodeAgent then rewrites \texttt{calculate\_objectives} rather than tuning scalar weights by hand. Table~\ref{tab:objective_learning_main} summarizes how distilled intent becomes code; Table~\ref{tab:objective_trajectory_kinase} links each revision to round-level hit-rate and docking gains.

\begin{table}[t]
  \caption{How LLM review is operationalized into training objectives (Surrogate-Expert--Agent, kinase FAK1, seed~0). Appendix~\ref{sec:appendix_objectives} lists full source code.}
  \label{tab:objective_learning_main}
  \centering
  \small
  \setlength{\tabcolsep}{3pt}
  \begin{tabular}{l p{0.28\linewidth} p{0.46\linewidth}}
    \toprule
    Obj. ver. & Intent distilled from review & Executable change \\
    \midrule
    0 (round 1) & Balance drug-likeness and diversity & Shared composite surrogate: QED, SA, Lipinski, novelty, batch diversity. \\
    2 (round 2) & Reject reactive warheads and PAINS & First LLM edit: PAINS filter, reactive-warhead hard SMARTS, hinge-binding rewards. \\
    3 (round 3) & Favor permeable, hinge-binding cores & Adds TPSA band for permeability; expands hinge/scaffold reward SMARTS. \\
    6 (round 6) & Refine ADME and kinase pharmacophores & Refines weights (QED, logP, TPSA, Fsp$^3$, SA); expands toxicophore filters. \\
    \bottomrule
  \end{tabular}
\end{table}

The objective begins as a compact composite (v0). By v2, the agent encodes chemist-style constraints directly: PAINS rejection, hard filters for Michael acceptors and reactive halides, and SMARTS rewards for hinge-binding motifs. Later versions add permeability shaping (TPSA), logP and Fsp$^3$ terms, and additional toxicophore filters---changes that coincide with the largest hit-rate jump at round~5 (25\% $\rightarrow$ 33\%). Full \texttt{objective\_v0}--\texttt{v6} source appears in Appendix~\ref{sec:appendix_objectives}.

On nuclear receptor (MCR), Surrogate-Expert--Agent does not sustain these gains (Table~\ref{tab:objective_trajectory_mcr}). Hit rate often peaks mid-optimization---27\% at round~3 on seed~0---then declines to 14\% by round~6 as later objective versions add progressively stricter developability rules: hard filters on anionic groups (v1), broad reactive-warhead SMARTS (v2), and then a formal-charge ban plus MW and Fsp$^3$ floors (v3). This pattern is consistent with a surrogate mismatch: the LLM encodes generic oral-drug heuristics that conflict with the charged, polar chemotypes common among MCR actives (${\sim}40\%$ of training SMILES carry explicit charge), whereas binary feedback---which directly rewards Tanimoto similarity to accepted actives, including charged hits---achieves the highest final-round hit rate on this target (21.2\%). MCR also has the lowest absolute hit rates across targets (${\sim}17$--21\%), yielding sparser docking signal per round and making feedback noisier; the combination of weak discovery signal and misaligned developability constraints likely explains why Surrogate-Expert--Agent trails simpler feedback baselines here despite competitive performance on kinase.

In the best-round view (Tables~\ref{tab:best_round_kinase}--\ref{tab:best_round_nuclear_receptor}, Appendix~\ref{sec:appendix_dude_tables}), peak hit rates reach 35--37\% for scalar and binary feedback on kinase, with similar cross-target patterns to the final-round tables. Per-target paired comparisons (Tables~\ref{tab:paired_deltas_kinase}--\ref{tab:paired_deltas_nuclear_receptor}, Appendix~\ref{sec:appendix_dude_tables}) reinforce this heterogeneity: on kinase, Surrogate-Expert--Agent improves over LVSEF-only by +4.4 percentage points on final-round hit rate; on GPCR, the gain is +2.0~pp; on MCR, Surrogate-Expert--Agent is within noise of LVSEF-only (+0.4~pp) and trails binary feedback ($-2.8$~pp). Agent-Agent warm-started from Surrogate-Expert--Agent knowledge is strongest on GPCR at both final and best rounds, suggesting that accumulated expert knowledge can support autonomous continuation when the target family permits it.

\textbf{Computational efficiency.} LVSEF-only trains in 0.55~h per run versus 8.81~h for DEG (${\sim}16\times$), with feedback conditions requiring ${\sim}4.2$~h per run due to six retraining rounds (full timing in Table~\ref{tab:training_timing}, Appendix~\ref{sec:appendix_dude_tables}).

Across docking and developability metrics, all conditions produce valid, diverse candidates on every target. Table~\ref{tab:objective_final_summary} summarizes final objectives per condition and target; round-by-round objective trajectories for kinase and MCR are in Tables~\ref{tab:objective_trajectory_kinase} and~\ref{tab:objective_trajectory_mcr} (Appendix~\ref{sec:appendix_objectives}).

\subsection{Prospective Laboratory Deployment}
\label{sec:results_deployment}

End-to-end results on the proprietary internal dataset are summarized in Table~\ref{end_to_end_results}. Among Human-Human, Human-Agent, and Agent-Agent configurations, Human-Agent performed best, identifying 13 molecules with docking scores below $-6$ compared with 7 for Human-Human---an 86\% improvement in hit identification. A collaborating medicinal chemist who conducted a blinded review ranked the Human-Agent output as the most promising overall.

These results suggest that the agentic system can operationalize expert feedback more effectively than a traditional human-to-human workflow, likely by reducing communication bottlenecks between medicinal chemists and AI engineers. Participating chemists reported that interacting with the system felt like ``talking to a colleague,'' indicating that the framework preserves natural expert communication while avoiding interpretation errors from human intermediaries.

Agent-Agent achieved 11 favorable docking hits, slightly below Human-Agent but with fully autonomous operation. The near-comparable performance combined with continuous availability makes Agent-Agent attractive when expert time is limited. The remaining gap may reflect current limitations of the surrogate medicinal-chemist agent, which reviews SMILES-level representations without 3D structural context or live literature retrieval.

\textbf{Limitations.} DUD-E Surrogate-Expert--Agent feedback is LLM-simulated rather than collected from the deployment chemists; we reserve Human-Agent for the prospective laboratory study. Final-round differences among feedback conditions are often within overlapping confidence intervals at five seeds per target; we therefore emphasize per-target trajectories, best-round peaks, and paired deltas rather than declaring a single dominant method across families. Docking scores are optimization signals and evaluation proxies, not experimental affinities.

\section{Conclusion and Future Work}

In drug lead optimization, it is common to have fewer than about 100 training molecules. Existing work addresses this setting with fragment-based models, but the selected fragments are often suboptimal for the generation process. It is also important to incorporate human domain experts to compensate for the lack of data, yet existing agentic models typically do not take raw human feedback. We introduced FRAGMENTA, an efficient end-to-end generative framework that addresses lead optimization in the small-data regime through an efficient fragment-based generative model and a closed-loop feedback agentic system. Our public benchmark experiments demonstrate that LVSEF outperforms state-of-the-art methods in extremely small-data regimes ($<$100 molecules) and remains competitive on datasets of roughly 100 molecules while training ${\sim}16\times$ faster. Our agentic system improves LVSEF over time by translating nuanced feedback into objective functions, raising final-round discovery yield by up to ${\sim}16\%$ relative over one-shot LVSEF-only on kinase when feedback quality is high. Our detailed analysis shows that these gains are target-dependent and largest when distilled knowledge is well matched to the chemistry, highlighting the framework's ability to incorporate expert feedback into generative optimization. Our real-world deployment in a cancer drug discovery campaign also demonstrated that the system identifies nearly twice as many molecules with favorable docking scores (docking score $< -6$) as baseline methods.

Future work will focus on extending the framework to larger molecular spaces and incorporating 3D structural information. We also plan to enhance the agentic reasoning engine by integrating real-time retrieval-augmented generation (RAG) from the scientific literature to further ground autonomous discovery in the latest medicinal chemistry insights.

\bibliographystyle{ACM-Reference-Format}
\bibliography{references}
\appendix
\section{Appendix}

\subsection{Additional DUD-E Result Tables}
\label{sec:appendix_dude_tables}

This appendix reports supporting DUD-E views used in the analysis: round-trajectory and per-target hit-rate plots (redundant with the main-text tables), best-round performance per target, Surrogate-Expert--Agent paired improvements versus automated baselines, and generative-model training times. Final-round tables appear in the main paper.

\subsubsection{Hit-rate trajectory and per-target views}
\begin{figure}[t]
  \centering
  \IfFileExists{data/figures/trajectory_hit_rate.png}{%
    \includegraphics[width=\linewidth]{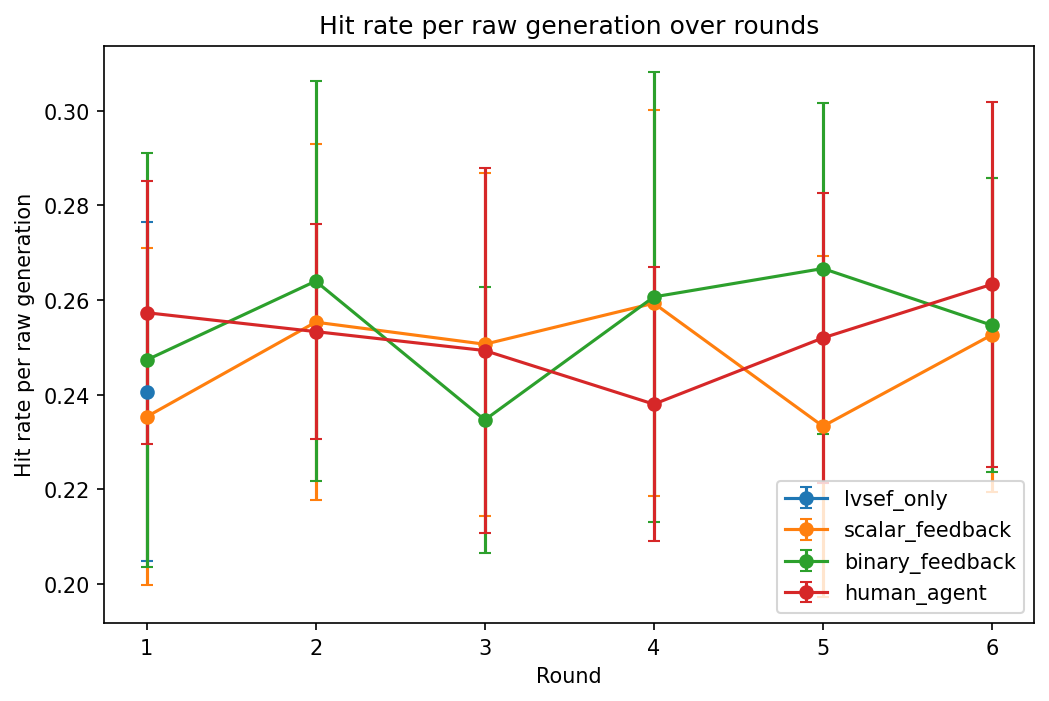}%
  }{%
    \fbox{\parbox{0.9\linewidth}{\centering Run \texttt{sync\_data.sh} to copy benchmark figures.}}%
  }
  \caption{Hit rate per raw generation over feedback rounds (mean $\pm$ 95\% CI). One-shot baselines (DEG, LVSEF-only) appear as horizontal references at round~1.}
  \label{fig:trajectory_hit}
\end{figure}

\begin{figure}[t]
  \centering
  \IfFileExists{data/figures/final_hit_rate_by_target.png}{%
    \includegraphics[width=\linewidth]{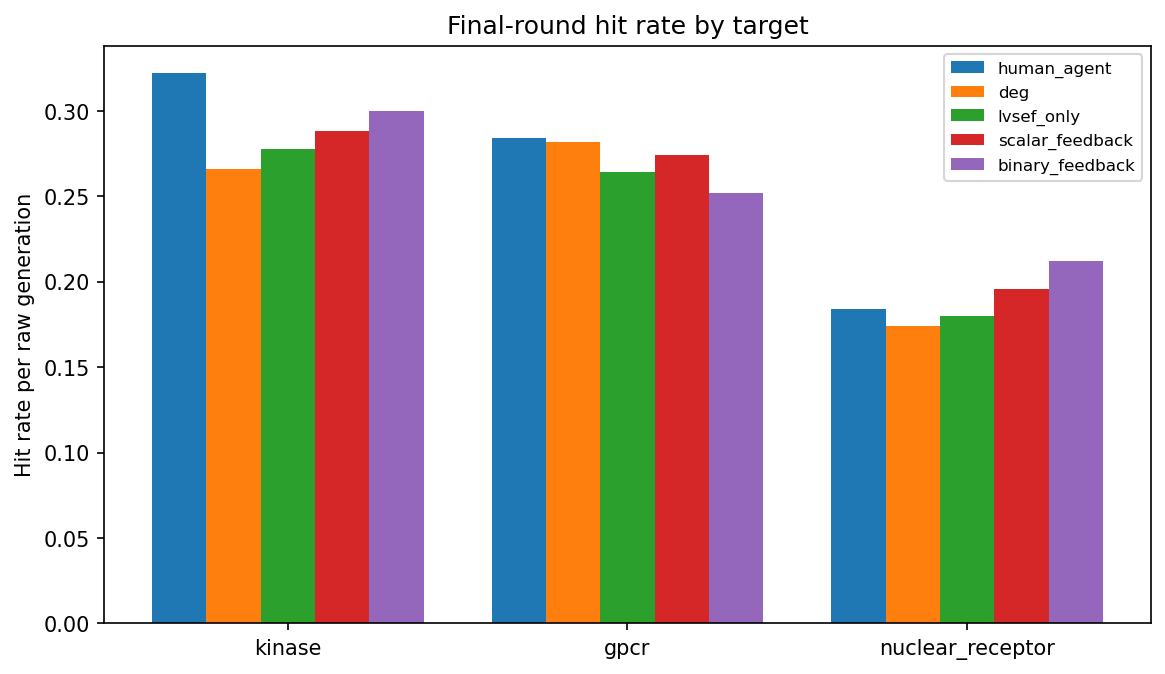}%
  }{%
    \fbox{\parbox{0.9\linewidth}{\centering Run \texttt{sync\_data.sh} to copy benchmark figures.}}%
  }
  \caption{Final-round hit rate by protein target and condition.}
  \label{fig:final_by_target}
\end{figure}

\subsubsection{Best-round performance}
\begin{table*}[t]
  \caption{Best-round performance on Kinase (FAK1) (argmax pre-registered selection score over rounds 1--6). Five seeds ($n{=}5$). Mean $\pm$ 95\% CI half-width.}
  \label{tab:best_round_kinase}
  \centering
  \footnotesize
  \setlength{\tabcolsep}{3.5pt}
  \begin{tabular}{l r r r r r r r r}
    \toprule
    Condition & Hit rate (\%) $\uparrow$ & Hits ($\leq -6$) $\uparrow$ & Top-10 Vina $\downarrow$ & Best Vina $\downarrow$ & QED $\uparrow$ & SA $\downarrow$ & Valid (\%) $\uparrow$ & Unique (\%) $\uparrow$ \\
    \midrule
    Surrogate-Expert--Agent & 33.6 ± 3.2\% & 33.6 ± 3.2 & -8.29 ± 0.39 & -9.75 ± 0.76 & 0.541 ± 0.019 & 0.538 ± 0.043 & 100.0 & \underline{99.8 $\pm$ 0.6} \\
    Agent-Agent & 32.6 ± 4.4\% & 32.6 ± 4.4 & \textbf{-8.79 ± 0.49} & \textbf{-10.15 ± 0.23} & 0.533 ± 0.016 & \underline{0.533 ± 0.029} & 100.0 & \underline{99.8 $\pm$ 0.6} \\
    Binary feedback & \textbf{37.4 ± 5.2\%} & \textbf{37.4 ± 5.2} & \underline{-8.61 ± 0.67} & -9.70 ± 1.11 & \underline{0.541 ± 0.018} & 0.553 ± 0.033 & 100.0 & 99.6 $\pm$ 1.1 \\
    Scalar feedback & \underline{35.8 ± 4.2\%} & \underline{35.8 ± 4.2} & -8.53 ± 0.51 & \underline{-10.13 ± 0.68} & \textbf{0.556 ± 0.028} & 0.557 ± 0.025 & 100.0 & 99.2 $\pm$ 1.0 \\
    DEG & 26.6 ± 6.5\% & 26.6 ± 6.5 & -8.13 ± 0.36 & -9.51 ± 0.99 & 0.502 ± 0.019 & \textbf{0.513 ± 0.028} & 100.0 & \textbf{100.0 $\pm$ 0.0} \\
    LVSEF-only & 27.8 ± 7.6\% & 27.8 ± 7.6 & -8.30 ± 0.21 & -9.71 ± 1.35 & 0.537 ± 0.031 & 0.535 ± 0.031 & 100.0 & \textbf{100.0 $\pm$ 0.0} \\
    \bottomrule
  \end{tabular}
\end{table*}

\begin{table*}[t]
  \caption{Best-round performance on GPCR (CXCR4) (argmax pre-registered selection score over rounds 1--6). Five seeds ($n{=}5$). Mean $\pm$ 95\% CI half-width.}
  \label{tab:best_round_gpcr}
  \centering
  \footnotesize
  \setlength{\tabcolsep}{3.5pt}
  \begin{tabular}{l r r r r r r r r}
    \toprule
    Condition & Hit rate (\%) $\uparrow$ & Hits ($\leq -6$) $\uparrow$ & Top-10 Vina $\downarrow$ & Best Vina $\downarrow$ & QED $\uparrow$ & SA $\downarrow$ & Valid (\%) $\uparrow$ & Unique (\%) $\uparrow$ \\
    \midrule
    Surrogate-Expert--Agent & 32.0 ± 2.3\% & 32.0 ± 2.3 & -8.27 ± 0.14 & -9.07 ± 0.50 & 0.505 ± 0.026 & 0.526 ± 0.015 & 100.0 & \underline{99.4 $\pm$ 1.1} \\
    Agent-Agent & \textbf{34.0 ± 2.3\%} & \textbf{34.0 ± 2.3} & \textbf{-8.64 ± 0.21} & \textbf{-9.98 ± 0.23} & \textbf{0.520 ± 0.021} & 0.532 ± 0.025 & 100.0 & 98.4 $\pm$ 1.9 \\
    Binary feedback & 32.2 ± 3.3\% & 32.2 ± 3.3 & -8.29 ± 0.27 & -9.42 ± 0.49 & 0.502 ± 0.026 & \underline{0.512 ± 0.030} & 100.0 & \textbf{99.6 $\pm$ 0.7} \\
    Scalar feedback & \underline{33.4 ± 3.2\%} & \underline{33.4 ± 3.2} & -8.35 ± 0.45 & -9.52 ± 0.74 & \underline{0.512 ± 0.024} & 0.518 ± 0.033 & 100.0 & 99.2 $\pm$ 1.6 \\
    DEG & 28.2 ± 6.3\% & 28.2 ± 6.3 & \underline{-8.43 ± 0.41} & \underline{-9.60 ± 0.35} & 0.507 ± 0.010 & 0.523 ± 0.024 & 100.0 & 99.0 $\pm$ 1.8 \\
    LVSEF-only & 26.4 ± 5.3\% & 26.4 ± 5.3 & -8.15 ± 0.36 & -9.09 ± 0.48 & 0.488 ± 0.011 & \textbf{0.504 ± 0.019} & 100.0 & \textbf{99.6 $\pm$ 0.7} \\
    \bottomrule
  \end{tabular}
\end{table*}

\begin{table*}[t]
  \caption{Best-round performance on Nuclear receptor (MCR) (argmax pre-registered selection score over rounds 1--6). Five seeds ($n{=}5$). Mean $\pm$ 95\% CI half-width.}
  \label{tab:best_round_nuclear_receptor}
  \centering
  \footnotesize
  \setlength{\tabcolsep}{3.5pt}
  \begin{tabular}{l r r r r r r r r}
    \toprule
    Condition & Hit rate (\%) $\uparrow$ & Hits ($\leq -6$) $\uparrow$ & Top-10 Vina $\downarrow$ & Best Vina $\downarrow$ & QED $\uparrow$ & SA $\downarrow$ & Valid (\%) $\uparrow$ & Unique (\%) $\uparrow$ \\
    \midrule
    Surrogate-Expert--Agent & \underline{24.8 ± 3.2\%} & \underline{24.8 ± 3.2} & \textbf{-8.12 ± 0.30} & \underline{-9.52 ± 1.22} & \underline{0.541 ± 0.011} & 0.527 ± 0.033 & 100.0 & \underline{99.8 $\pm$ 0.6} \\
    Agent-Agent & 23.6 ± 3.2\% & 23.6 ± 3.2 & -7.87 ± 0.52 & -9.00 ± 1.05 & 0.540 ± 0.015 & 0.525 ± 0.018 & 100.0 & \textbf{100.0 $\pm$ 0.0} \\
    Binary feedback & \textbf{25.6 ± 1.9\%} & \textbf{25.6 ± 1.9} & \underline{-8.02 ± 0.27} & -9.23 ± 0.57 & 0.534 ± 0.026 & 0.527 ± 0.016 & 100.0 & \textbf{100.0 $\pm$ 0.0} \\
    Scalar feedback & 23.8 ± 4.2\% & 23.8 ± 4.2 & -7.76 ± 0.32 & -8.62 ± 0.53 & \textbf{0.548 ± 0.009} & 0.528 ± 0.039 & 100.0 & 99.6 $\pm$ 0.7 \\
    DEG & 17.4 ± 2.4\% & 17.4 ± 2.4 & -7.42 ± 0.19 & -8.74 ± 0.68 & 0.520 ± 0.007 & \underline{0.520 ± 0.026} & 100.0 & \textbf{100.0 $\pm$ 0.0} \\
    LVSEF-only & 18.0 ± 5.6\% & 18.0 ± 5.6 & -7.74 ± 0.40 & \textbf{-9.61 ± 1.60} & 0.513 ± 0.023 & \textbf{0.509 ± 0.023} & 100.0 & \underline{99.8 $\pm$ 0.6} \\
    \bottomrule
  \end{tabular}
\end{table*}

\subsubsection{Paired improvement (Surrogate-Expert--Agent vs.\ baselines)}
\begin{table}[t]
  \caption{Surrogate-Expert--Agent paired improvement vs.\ automated baselines on Kinase (FAK1) (final-round view, $n{=}5$ matched seeds). $\Delta$ = Surrogate-Expert--Agent minus baseline; for docking scores, $\Delta$ = baseline minus Surrogate-Expert--Agent (positive = stronger binding).}
  \label{tab:paired_deltas_kinase}
  \centering
  \small
  \setlength{\tabcolsep}{4pt}
  \begin{tabular}{l r r r r}
    \toprule
    Comparison & $\Delta$ hit rate (\%) & $\Delta$ hits & $\Delta$ top-10 Vina & $\Delta$ QED \\
    \midrule
    vs.\ LVSEF-only & +4.4 ± 5.2\% & +4.4 ± 5.2 & +0.13 ± 0.32 & +0.004 ± 0.029 \\
    vs.\ scalar feedback & +3.4 ± 2.4\% & +3.4 ± 2.4 & -0.03 ± 0.33 & +0.020 ± 0.015 \\
    vs.\ binary feedback & +2.2 ± 8.8\% & +2.2 ± 8.8 & +0.06 ± 0.32 & +0.006 ± 0.011 \\
    \bottomrule
  \end{tabular}
\end{table}

\begin{table}[t]
  \caption{Surrogate-Expert--Agent paired improvement vs.\ automated baselines on GPCR (CXCR4) (final-round view, $n{=}5$ matched seeds). $\Delta$ = Surrogate-Expert--Agent minus baseline; for docking scores, $\Delta$ = baseline minus Surrogate-Expert--Agent (positive = stronger binding).}
  \label{tab:paired_deltas_gpcr}
  \centering
  \small
  \setlength{\tabcolsep}{4pt}
  \begin{tabular}{l r r r r}
    \toprule
    Comparison & $\Delta$ hit rate (\%) & $\Delta$ hits & $\Delta$ top-10 Vina & $\Delta$ QED \\
    \midrule
    vs.\ LVSEF-only & +2.0 ± 5.6\% & +2.0 ± 5.6 & -0.10 ± 0.31 & +0.014 ± 0.012 \\
    vs.\ scalar feedback & +1.0 ± 4.2\% & +1.0 ± 4.2 & -0.12 ± 0.23 & +0.006 ± 0.023 \\
    vs.\ binary feedback & +3.2 ± 4.2\% & +3.2 ± 4.2 & +0.06 ± 0.10 & +0.004 ± 0.027 \\
    \bottomrule
  \end{tabular}
\end{table}

\begin{table}[t]
  \caption{Surrogate-Expert--Agent paired improvement vs.\ automated baselines on Nuclear receptor (MCR) (final-round view, $n{=}5$ matched seeds). $\Delta$ = Surrogate-Expert--Agent minus baseline; for docking scores, $\Delta$ = baseline minus Surrogate-Expert--Agent (positive = stronger binding).}
  \label{tab:paired_deltas_nuclear_receptor}
  \centering
  \small
  \setlength{\tabcolsep}{4pt}
  \begin{tabular}{l r r r r}
    \toprule
    Comparison & $\Delta$ hit rate (\%) & $\Delta$ hits & $\Delta$ top-10 Vina & $\Delta$ QED \\
    \midrule
    vs.\ LVSEF-only & +0.4 ± 4.4\% & +0.4 ± 4.4 & -0.08 ± 0.34 & +0.006 ± 0.022 \\
    vs.\ scalar feedback & -1.2 ± 4.6\% & -1.2 ± 4.6 & -0.18 ± 0.24 & -0.016 ± 0.022 \\
    vs.\ binary feedback & -2.8 ± 4.4\% & -2.8 ± 4.4 & +0.02 ± 0.25 & -0.002 ± 0.005 \\
    \bottomrule
  \end{tabular}
\end{table}

\subsubsection{Training cost}
\begin{table}[t]
  \caption{Generative-model training cost per benchmark run (LVSEF/DEG \texttt{train} + \texttt{evaluate} subprocess wall time). Excludes Vina docking, LLM calls, and feedback logic. Mean $\pm$ 95\% CI half-width over 15 runs per condition.}
  \label{tab:training_timing}
  \centering
  \small
  \begin{tabular}{l r r}
    \toprule
    Condition & Train (h) & Train + eval (h) \\
    \midrule
    LVSEF-only      & \textbf{0.55 $\pm$ 0.12} & \textbf{0.72 $\pm$ 0.17} \\
    Surrogate-Expert--Agent     & 3.27 $\pm$ 0.76 & 4.25 $\pm$ 1.02 \\
    Scalar feedback & 3.27 $\pm$ 0.77 & 4.23 $\pm$ 1.01 \\
    Binary feedback & 3.28 $\pm$ 0.77 & 4.24 $\pm$ 1.02 \\
    Agent-Agent     & 3.25 $\pm$ 0.75 & 4.22 $\pm$ 1.00 \\
    DEG             & 8.81 $\pm$ 4.28 & 8.97 $\pm$ 4.29 \\
    \bottomrule
  \end{tabular}
\end{table}

\subsection{Experiment Protocol Details}
\label{sec:appendix_experiment_protocol}

\subsubsection{Small-data molecular generation}
\paragraph{Datasets.}
We evaluate on Chain Extenders (11 molecules)\cite{guo2022data}, Acrylates (32 molecules)\cite{guo2022data}, and a proprietary cancer-lead set (104 experimentally validated actives). These span extreme small data to a more typical lead-optimization scale (${\sim}$100 molecules).

\paragraph{Baselines.}
We compare against GraphNVP \cite{madhawa2019graphnvp} (atom-based), HierVAE \cite{jin2020hierarchical} and MiCaM \cite{geng2023novo} (fragment-based with static or deep-learned vocabularies), and DEG \cite{guo2022data} (small-data SOTA fragment grammar). Data-hungry diffusion and large deep generative models are omitted because they fail to generalize at these scales.

\paragraph{LVSEF variants and objectives.}
\underline{LVSEF(ran)} samples among top-ranked fragments; \underline{LVSEF(bal)} samples probabilistically to balance exploration and exploitation. Primary objectives on Chain Extenders and Acrylates are synthesizability and diversity \cite{gao2021amortized,bradshaw2020barking,gao2024generative}. On the internal set, training penalizes MW $>500$ or $\log P>5$.

\paragraph{Molecule-generation metrics.}
Beyond uniqueness, novelty, and validity \cite{brown2019guacamol,geng2023novo}, we report diversity (mean pairwise Tanimoto distance over Morgan fingerprints)\cite{bajusz2015tanimoto}, Chamfer distance to the training set\cite{fan2017point}, QED\cite{bickerton2012quantifying}, Lipinski compliance\cite{lipinski2012experimental}, SA (lower is better)\cite{ertl2009estimation}, Retro Score (RS)\cite{chen2020retro}, scaffold diversity (unique Bemis--Murcko scaffolds), membership (fraction in the target monomer class), and discovery rate (unique, novel, synthesizable molecules, with and without membership). We generate 1{,}000 molecules for Chain Extenders and Acrylates; for the internal set we generate 10{,}000 and retain the top 100 by internal scoring.

\subsubsection{DUD-E lead optimization}
\paragraph{Targets.}
We evaluate one representative target from each of three protein families under a frozen selection policy (80--200 training actives, receptor availability, and successful docking verification):
\begin{itemize}
    \item \textbf{Kinase (FAK1):} Focal adhesion kinase 1 (\texttt{fak1}); 113 training actives.
    \item \textbf{GPCR (CXCR4):} C-X-C chemokine receptor type 4 (\texttt{cxcr4}); 118 training actives.
    \item \textbf{Nuclear receptor (MCR):} Mineralocorticoid receptor (\texttt{mcr}); 185 training actives.
\end{itemize}

\paragraph{Baselines and budget.}
Each feedback round generates 100 molecules \cite{guo2022data}. One-shot references (DEG, LVSEF-only) train for 10 epochs once; feedback methods run six rounds with 100 molecules per round.
\begin{itemize}
    \item \textbf{DEG} \cite{guo2022data}: data-efficient fragment grammar with randomly combined fragments.
    \item \textbf{LVSEF-only:} fixed initial objective; no closed-loop tuning.
    \item \textbf{Scalar feedback:} updates the objective from continuous composite scores on top-ranked candidates.
    \item \textbf{Binary feedback:} accepts/rejects candidates by a fixed score threshold before updating the objective.
    \item \textbf{Surrogate-Expert--Agent:} LLM medicinal-chemist surrogate drives objective updates via the agentic stack (same interface as Human-Agent in Figure~\ref{fragmenta_overview}b; we reserve Human-Agent for real chemists in deployment).
    \item \textbf{Agent-Agent:} autonomous continuation of Surrogate-Expert--Agent for the same target, warm-started from accumulated knowledge.
\end{itemize}
Human-Human is evaluated only in the prospective deployment (Section~\ref{sec:deployment}), where repeated expert interaction across three public targets, five seeds, and six rounds is impractical.

\paragraph{DUD-E metrics.}
Our primary metric is \textbf{hit rate (raw generation)}: the fraction of the 100 raw generated molecules with AutoDock Vina score $< -6$ kcal/mol (discovery yield under a fixed budget). We additionally report docking hit count; best and top-10 mean Vina scores; validity and uniqueness; QED; SA (lower is better); Lipinski compliance; novelty ($1-\max$ Tanimoto to training actives); pairwise Tanimoto diversity; and Bemis--Murcko scaffold count. Best-round and total-budget views appear in Appendix~\ref{sec:appendix_dude_tables}. Mean docking over all molecules is not a headline metric because invalid generations are excluded from docking.

\subsubsection{Prospective deployment details}
\label{sec:appendix_deployment}
FRAGMENTA ran as a command-line tool integrated into chemists' existing workflows. Each week the system ranked candidates and presented the top 100 for review. The campaign covered six complete feedback rounds over six months, aiming to identify molecules with promising docking scores for subsequent in vitro validation. Two medicinal chemists participated throughout: a postdoctoral researcher as primary weekly reviewer and a professor-level supervisor for strategic guidance. Both chemists remained consistent, ensuring continuity of evaluation criteria. This study uses real medicinal chemists (Human-Human and Human-Agent in Figure~\ref{fragmenta_overview}) and does not use the LLM surrogate from the public benchmark.

\subsection{Computing Setup}

Our experiments were conducted on a high-performance computing cluster consisting of 32 compute nodes. Each node features dual AMD EPYC 7502 32-core processors (64 cores per node), 512GB DDR4 memory, and dual 960GB SSDs, providing a total of 2,048 cores, 16TB memory, and distributed storage across the cluster. The nodes are organized in 8 rack-mount drawers with 4 nodes per drawer, and we utilized 10 cores from this infrastructure for our experiments.

\subsection{Computational Efficiency and Cost Analysis}

The deployment of FRAGMENTA demonstrated a significant reduction in the temporal bottleneck traditionally associated with expert-in-the-loop molecular design. We evaluate the system's efficiency across two primary dimensions: Communication Latency and Operational Cost.

\subsubsection{Temporal Efficiency: Communication vs. Training}

In the traditional Human-Human mode, a single tuning cycle typically spanned approximately two weeks, including 1) Asynchronous Communication (1 week), where email exchanges and in-person meetings between the medicinal chemist and AI engineer were held to clarify intent and translate intuition into code, and 2) Model Execution (1 week), that is the computational time required for training and candidate generation using the LVSEF.

By deploying the Human-Agent mode, the communication phase is shortened from one week to negligible intervals. Each agentic interaction (Perception $\rightarrow$ Extraction $\rightarrow$ Implementation) is processed within minutes. Given that a comprehensive tuning cycle involves approximately 5–10 such interactions, the agentic framework effectively reduces the end-to-end design cycle to the duration of the model training itself (approx. 1 week), representing a 50\% improvement in the cycle speed in total.

\subsubsection{Operational and API Cost Analysis}

To ensure reproducibility and assess the economic feasibility of the system for laboratory deployment, we tracked the token usage of our LLM-based controller. We utilized the Gemini 2.5 Pro API as the core reasoning engine for our agents. Below we report our observations:

\begin{itemize}
    \item \textit{Per-Turn Cost}: A single turn (a question-answer pair between the chemist and the agent) involves multiple LLM calls for validation, abstraction, and code generation.
    \item \textit{Total Iteration Cost}: One full iteration, encompassing the complete multi-agent feedback loop, averaged approximately \$0.10 USD.
    \item \textit{Total Project Cost}: Across the seven-month deployment, the total API expenditure was negligible relative to the cost of standard wet-lab reagents or the time of a postdoctoral researcher.
\end{itemize}

\subsection{Ablation Study and Sensitivity Analyses for LVSEF}

We have performed a number of experiments to extend ablation study of LVSEF as follows.

\begin{figure}[tp]
    \centering
    \includegraphics[width=0.85\columnwidth]{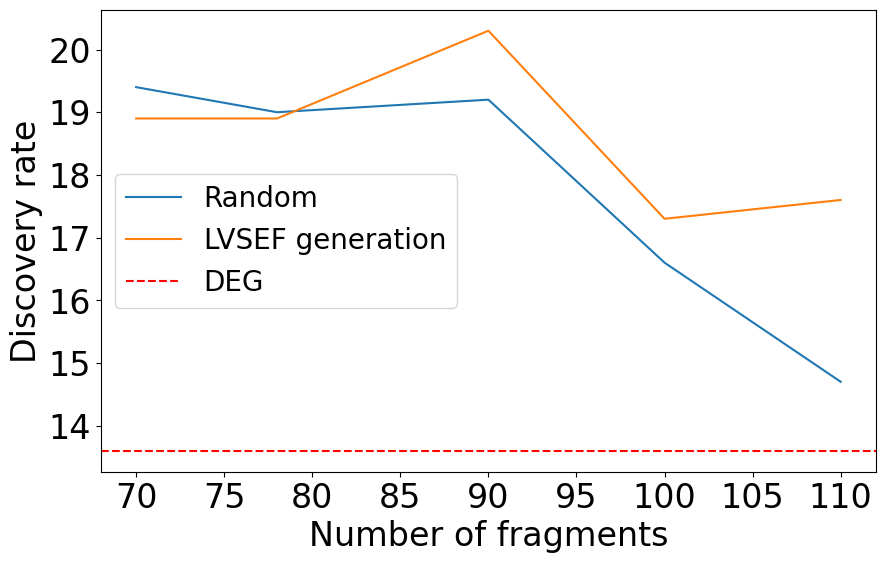}
    \caption{Ablation Study on Generative Architectures: Comparing the uniqueness and validity of molecules generated by the LVSEF framework against baseline configurations.}
    \label{fig:ablation_generation}
\end{figure}

\begin{figure}[tp]
    \centering
    \includegraphics[width=0.85\columnwidth]{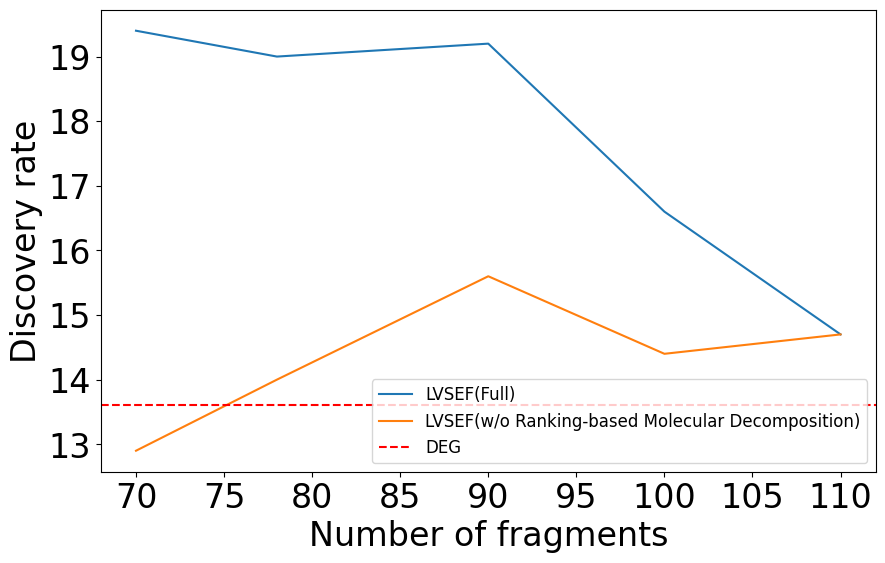}
    \caption{Ablation Study on Molecular Fragment Ranking (MFR): Evaluating the impact of ranking-based decomposition on the discovery rate of high-utility fragments.}
    \label{fig:ablation_mfr}
\end{figure}

\begin{figure}[tp]
    \centering
    \includegraphics[width=0.85\columnwidth]{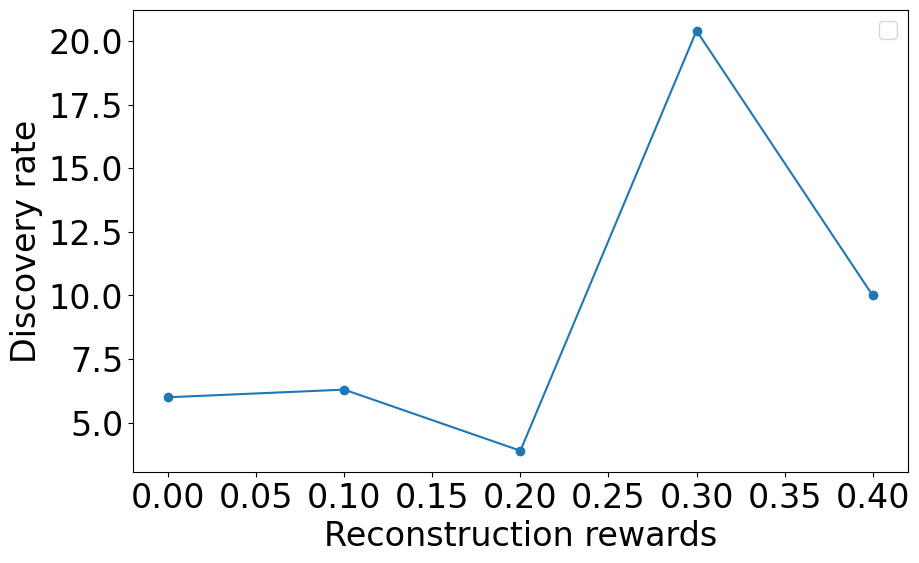}
    \caption{Ablation Study on Reconstruction Rewards ($R_{rec}$): Investigating how additive rewards for training-set connections ground the model in realistic chemical space.}
    \label{fig:ablation_reconstruction}
\end{figure}

\begin{figure}[tp] 
    \centering
    \includegraphics[width=0.85\columnwidth]{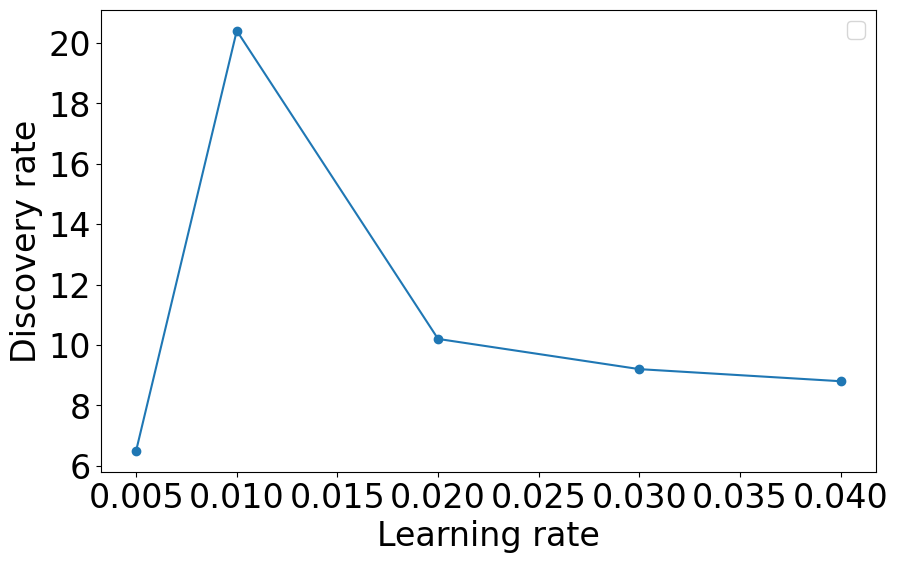} 
    \caption{Ablation Study on Learning Rate ($\alpha$): Investigating the impact of the learning rate on the convergence of Q-values within the fragment connection space.}
    \label{fig:learning_rate}
\end{figure}

\begin{figure*}[h]
\centering
\includegraphics[scale=0.2]{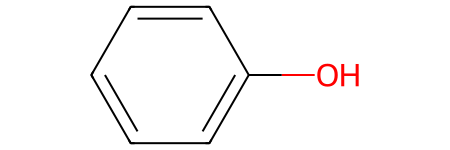}
\hspace{3mm} 
\includegraphics[scale=0.2]{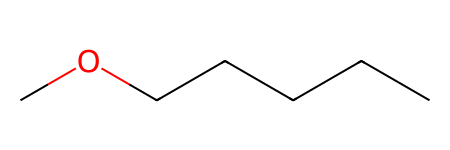}
\hspace{3mm} 
\includegraphics[scale=0.2]{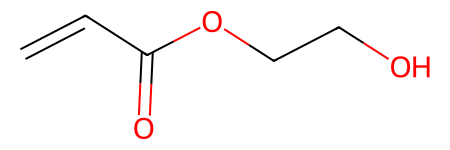}
\caption{Sample result: Fragment Samples (Left and Center), and LVSEF-Generated sample (Right)}
\label{sample}
\end{figure*}

\paragraph{Generation Strategy}
Here we evaluate the efficacy of our novel molecule generation strategy, which employs a Markov chain model enhanced by a learned Q-table. The study results are shown in Figure \ref{fig:ablation_generation}, where we compare the performance of our method against the random combination technique (where the likelihood of adding closed fragments is increased over time to avoid endless connections) that is used in prior research by \citep{guo2022data} using Acrylates dataset. The results demonstrate that our generation strategy outperforms the random combination method especially when we select more fragments that allow for synthesis of more combinations. Furthermore, we utilize this generation process not only as a means of producing viable molecular structures but also as an integral part of our training methodology.

\paragraph{Molecular Fragment Ranking}

 With this ablation experiment, we aim to evaluate the effectiveness of our Molecular Fragment Ranking and Ranking-based Molecular Decomposition through a detailed comparison involving three distinct models: LVSEF; LVSEF (without Ranking-based Molecular Decomposition), which adopts a similar approach to \citep{guo2022data} using a Graph Neural Network (GNN) to predict cutting points rather than using our ranking-based method; and finally the SOTA method DEG. For a fair comparison, all models employ the same random combination for molecule generation. The results, shown in Figure \ref{fig:ablation_mfr}, demonstrate that Ranking-based Molecular Decomposition outperforms GNN-based decomposition. This is explained by the fact that GNN-based decomposition requires significant amount of training data to learn, and with limited datasets, our reinforcement learning-based fragment ranking outperforms this approach.

 \paragraph{Reconstruction Reward}

We also conducted a sensitivity analysis for the reconstruction reward to determine its impact on the model’s learning efficiency. The results depicted in Figure \ref{fig:ablation_reconstruction} indicate that the incorporation of reconstruction rewards significantly enhances the ability of the model to learn effectively. This mechanism aids the model in capturing the underlying data distribution more accurately, thereby improving the overall performance. However, the analysis also reveals that excessively high reconstruction rewards can lead to overfitting, underscoring the necessity of a balanced approach in setting the reward level. Thus, careful calibration of the reconstruction reward is crucial to maintain the model’s generalization capabilities while maximizing its learning efficiency. This finding highlights the delicate balance required in tuning hyperparameters to optimize model performance without compromising its robustness and adaptability to new data.

\paragraph{Effect of Learning Rate}

We conducted a sensitivity study on the learning rate of the Q-table using the Acrylates dataset, as shown in Figure \ref{fig:learning_rate}. The results indicate that the optimal discovery rate occurs at a learning rate of approximately 0.01, with performance deteriorating as the learning rate increases. This trend can be attributed to the fact that a very small learning rate results in slower learning progress, while an excessively high learning rate prevents the model from converging effectively, similar to what is observed in deep learning models.

\paragraph{Observations on Learned Fragments}

\begin{table}[t]
\large

\centering
\begin{tabular}{|l|l|}
\hline
\textbf{Model} & \textbf{Average number of atoms} \\
\hline
DEG & 3.0 \\
LVSEF & 4.2 \\
\hline
\end{tabular}
\caption{Average Number of Atoms in Fragments}
\label{num_atoms_fragments}
\end{table}

We further investigated the complexity of fragments generated by our model compared to those produced by the DEG model. Table \ref{num_atoms_fragments} presents the average number of atoms per fragment for both DEG and LVSEF when the number of fragments is equal (78 for Acrylates). This shows that our fragments contain more atoms, resulting in greater complexity. This increased complexity is achievable because our model can extract multiple fragments from a single atom within a molecule, whereas the state-of-the-art (SOTA) methods split a molecule into fragments, limiting their size. Consequently, the more complex fragments produced by our model enhance its ability to discover novel and synthesizable molecules.

\paragraph{Fragment and Generated Samples}

Finally, to demonstrate capability of our proposed model, we show a sample outcome. Figure \ref{sample} shows two fragment samples as well as a sample generated by LVSEF trained with Acrylates dataset. Those two fragments are phenol and 1-pentyne, which are known to be synthesizable. The LVSEF-generated sample includes Hydroxyethyl acrylate which is a well-known acrylates, which does not appear in DEG-generated samples.

\subsection{Implementation Details of the Agentic System}

\subsection{Benchmark Training Objectives}
\label{sec:appendix_objectives}

Each DUD-E feedback round retrains LVSEF with a Python \texttt{calculate\_objectives} function. Round~1 starts from a shared composite RDKit surrogate. Surrogate-Expert--Agent and Agent-Agent pass docking-ranked candidates through an LLM medicinal-chemist surrogate; the CodeAgent then rewrites the objective to encode the review as SMARTS filters, property bands, and reweighted scoring terms.

\begin{table*}[t]
  \caption{Final training objective per DUD-E condition and target (seed~0, highest \texttt{objective\_v*} after six feedback rounds).}
  \label{tab:objective_final_summary}
  \centering
  \footnotesize
  \setlength{\tabcolsep}{4pt}
  \begin{tabular}{l p{0.28\linewidth} p{0.28\linewidth} p{0.28\linewidth}}
    \toprule
    Condition & Kinase (FAK1) & GPCR (CXCR4) & Nuclear receptor (MCR) \\
    \midrule
    Surrogate-Expert--Agent & v13: PAINS filter; 45 SMARTS patterns; 14 hard filters; TPSA shaping; logP shaping; weighted property mix & v4: 3 SMARTS patterns & v5: PAINS filter; 3 SMARTS patterns \\
    Agent-Agent & v6: PAINS filter; 3 SMARTS patterns & v5: 4 SMARTS patterns & v6: PAINS filter; 3 SMARTS patterns; weighted property mix \\
    Scalar feedback & v1: alpha=0.3 centroid Tanimoto & v1: alpha=0.3 centroid Tanimoto & v1: alpha=0.3 centroid Tanimoto \\
    Binary feedback & v1: beta=0.4 accept-set Tanimoto & v1: beta=0.4 accept-set Tanimoto & v1: beta=0.4 accept-set Tanimoto \\
    LVSEF-only & v0: weighted QED/SA/Lipinski/novelty/diversity composite & v0: LLM-authored composite objective & v0: LLM-authored composite objective \\
    DEG & v0: LLM-authored composite objective & v0: LLM-authored composite objective & v0: LLM-authored composite objective \\
    \bottomrule
  \end{tabular}
\end{table*}

\subsubsection{Round-by-round learning on kinase (Surrogate-Expert--Agent, seed~0)}
\begin{table*}[t]
  \caption{Per-round objective updates and discovery gains for Surrogate-Expert--Agent on kinase (FAK1), seed~0. Each row is the objective active at the start of that round (v0 at round~1; v2--v6 after subsequent LLM/CodeAgent edits). $\Delta$ columns compare to the prior round.}
  \label{tab:objective_trajectory_kinase}
  \centering
  \footnotesize
  \setlength{\tabcolsep}{3pt}
  \begin{tabular}{r r p{0.38\linewidth} r r r r}
    \toprule
    Round & Ver. & LLM-driven objective change & Hit (\%) $\uparrow$ & $\Delta$ hit & Best Vina $\downarrow$ & $\Delta$ Vina \\
    \midrule
    1 & 0 & Shared composite surrogate: QED, SA, Lipinski, novelty, batch diversity. & 25.0 & --- & -8.90 & --- \\
    2 & 2 & First LLM edit: PAINS filter, reactive-warhead hard SMARTS, hinge-binding rewards. & 24.0 & -1.0 & -9.25 & -0.3 \\
    3 & 3 & Adds TPSA band for permeability; expands hinge/scaffold reward SMARTS. & 25.0 & +1.0 & -9.43 & -0.2 \\
    4 & 4 & Further kinase pharmacophore rewards (pyridine/pyrimidine hinges, bioisosteres). & 23.0 & -2.0 & -9.62 & -0.2 \\
    5 & 5 & Adds logP shaping and additional reward motifs from review feedback. & 33.0 & +10.0 & -10.05 & -0.4 \\
    6 & 6 & Refines weights (QED, logP, TPSA, Fsp$^3$, SA); expands toxicophore filters. & 32.0 & -1.0 & -10.08 & -0.0 \\
    \bottomrule
  \end{tabular}
\end{table*}

\subsubsection{Round-by-round learning on nuclear receptor (Surrogate-Expert--Agent, seed~0)}
\begin{table*}[t]
  \caption{Per-round objective updates and discovery gains for Surrogate-Expert--Agent on nuclear receptor (MCR), seed~0. Each row is the objective active at the start of that round (v0 at round~1; v1--v5 after subsequent LLM/CodeAgent edits). $\Delta$ columns compare to the prior round.}
  \label{tab:objective_trajectory_mcr}
  \centering
  \footnotesize
  \setlength{\tabcolsep}{3pt}
  \begin{tabular}{r r p{0.38\linewidth} r r r r}
    \toprule
    Round & Ver. & LLM-driven objective change & Hit (\%) $\uparrow$ & $\Delta$ hit & Best Vina $\downarrow$ & $\Delta$ Vina \\
    \midrule
    1 & 0 & Shared composite surrogate: QED, SA, Lipinski, novelty, batch diversity. & 19.0 & --- & -8.99 & --- \\
    2 & 1 & First LLM edit: hard filters on anionic O/N (permeability); thiophene/ester penalties. & 20.0 & +1.0 & -9.86 & -0.9 \\
    3 & 2 & Adds Michael-acceptor, acetal, and additional reactive-warhead hard SMARTS. & 27.0 & +7.0 & -9.71 & +0.1 \\
    4 & 3 & Rejects any formal charge; enforces MW 250--500 Da and Fsp$^3 \geq 0.2$; adds epoxide filter. & 17.0 & -10.0 & -10.15 & -0.4 \\
    5 & 4 & Adds pharmacophore reward SMARTS (phenol, sulfone, benzimidazole, fluorophenyl). & 16.0 & -1.0 & -8.75 & +1.4 \\
    6 & 5 & Expands pharmacophore bonuses and rotatable-bond / complexity limits. & 14.0 & -2.0 & -8.56 & +0.2 \\
    \bottomrule
  \end{tabular}
\end{table*}

\paragraph{Representative objective source (v0 and v2--v6).}
Full files are archived under \texttt{paper/data/objectives/benchmark/}. Below we show the initial composite (v0) and excerpts of later LLM-authored objectives that encode medicinal-chemistry constraints as SMARTS filters and weighted scoring terms.

\subsubsection*{Objective v0 (full)}
\begin{lstlisting}[language=Python,caption={\texttt{objective\_v0.py}}]
"""Cheap RDKit composite objective for LVSEF training (no Vina in inner loop)."""

from __future__ import annotations

try:
    from rdkit import Chem
    from rdkit.Chem import Descriptors, QED
    from rdkit import DataStructs
    from rdkit.Chem import AllChem
except ImportError:
    Chem = None

_TRAIN_FPS: list = []


def _load_train_fps(train_file: str | None = None) -> None:
    global _TRAIN_FPS
    if _TRAIN_FPS or Chem is None:
        return
    if train_file:
        from pathlib import Path

        p = Path(train_file)
        if p.exists():
            for line in p.read_text().splitlines():
                line = line.strip()
                if not line:
                    continue
                mol = Chem.MolFromSmiles(line)
                if mol:
                    _TRAIN_FPS.append(
                        AllChem.GetMorganFingerprintAsBitVect(mol, 2, nBits=2048)
                    )


def _sa_norm(mol) -> float:
    try:
        n_atoms = mol.GetNumAtoms()
        n_rings = len(Chem.GetSSSR(mol))
        sa = 1 + (n_atoms / 10) + (n_rings * 0.5)
        sa = min(10.0, max(1.0, sa))
        return max(0.0, min(1.0, 1.0 - (sa - 1.0) / 9.0))
    except Exception:
        return 0.5


def _lipinski_pass(mol) -> float:
    mw = Descriptors.MolWt(mol)
    logp = Descriptors.MolLogP(mol)
    hbd = Descriptors.NumHDonors(mol)
    hba = Descriptors.NumHAcceptors(mol)
    return 1.0 if mw <= 500 and logp <= 5 and hbd <= 5 and hba <= 10 else 0.0


def _novelty(mol) -> float:
    if not _TRAIN_FPS:
        return 1.0
    fp = AllChem.GetMorganFingerprintAsBitVect(mol, 2, nBits=2048)
    max_sim = max(DataStructs.TanimotoSimilarity(fp, t) for t in _TRAIN_FPS)
    return 1.0 - max_sim


def calculate_objectives(mol_list, diversity_calculator=None):
  # LVSEF passes mol objects or SMILES depending on version; handle both
    if Chem is None:
        return [0.5] * len(mol_list)
    scores = []
    batch_fps = []
    mols = []
    for item in mol_list:
        if isinstance(item, str):
            mol = Chem.MolFromSmiles(item)
        else:
            mol = item
        mols.append(mol)
        if mol:
            batch_fps.append(AllChem.GetMorganFingerprintAsBitVect(mol, 2, nBits=2048))
        else:
            batch_fps.append(None)

    for mol, fp in zip(mols, batch_fps):
        if mol is None or fp is None:
            scores.append(0.0)
            continue
        qed = QED.qed(mol)
        sa = _sa_norm(mol)
        lip = _lipinski_pass(mol)
        nov = _novelty(mol)
        div = 0.5
        if len(batch_fps) > 1:
            dists = []
            for other in batch_fps:
                if other is not None and other is not fp:
                    dists.append(1.0 - DataStructs.TanimotoSimilarity(fp, other))
            if dists:
                div = sum(dists) / len(dists)
        score = 0.35 * qed + 0.25 * sa + 0.15 * lip + 0.15 * nov + 0.10 * div
        scores.append(max(0.0, min(1.0, score)))
    return scores


_load_train_fps('train_actives.txt')  # path supplied at runtime
\end{lstlisting}

\subsubsection*{Objective v2 (excerpt)}
\begin{lstlisting}[language=Python,caption={\texttt{objective\_v2.py}}]
"""Excerpt: objective_v2.py (Human-Agent, kinase, seed 0)."""

_HARD_PENALTY_SUBSTRUCTURES = {
    # Penalize permanent charges which limit cell permeability.
    "charge": Chem.MolFromSmarts("[*;+1,*;+2,*;+3,*;-1,*;-2,*;-3]"),
    # Penalize PAINS-like quinone moieties, a common structural alert.
    "pains_quinone": Chem.MolFromSmarts("O=c1ccccc1=O"),
    # Penalize potentially reactive/unstable iminyl fluorides.
    "iminyl_fluoride": Chem.MolFromSmarts("[#6](=[#7])-[#9]"),

}

_SOFT_PENALTY_SUBSTRUCTURES = {
    # Penalize unsubstituted phenols, a metabolic liability.
    "unsubstituted_phenol": Chem.MolFromSmarts("[#6;a]-[#8;H1]"),

}

_REWARD_SUBSTRUCTURES = {
    # Validated hinge-binding motifs
    "pyrimidine": Chem.MolFromSmarts("c1ncncn1"),
    "thiadiazole": Chem.MolFromSmarts("c1csnn1"),
    # Modern medicinal chemistry motifs
    "sulfoximine": Chem.MolFromSmarts("[#16](=[#8])(=[#7])"),
    "alkyl_nitrile": Chem.MolFromSmarts("[#6]-[#6]#[#7]"),
    "trifluoromethyl": Chem.MolFromSmarts("C(F)(F)F"),

}

# Scoring combination
base_score = (
                0.25 * qed
                + 0.20 * sa
                + 0.10 * lip
                + 0.10 * nov
                + 0.15 * div
                + 0.20 * reward
            )

            # Step 4: Apply soft penalties as a multiplicative factor
            penalty_multiplier = _apply_soft_penalties(mol)
            final_score = base_score * penalty_multiplier

            # Step 5: Ensur
    ...
\end{lstlisting}

\subsubsection*{Objective v3 (excerpt)}
\begin{lstlisting}[language=Python,caption={\texttt{objective\_v3.py}}]
"""Excerpt: objective_v3.py (Human-Agent, kinase, seed 0)."""

_HARD_PENALTY_SUBSTRUCTURES = {
    # Penalize permanent charges which limit cell permeability.
    "charge": Chem.MolFromSmarts("[*;+1,*;+2,*;+3,*;-1,*;-2,*;-3]"),
    # Penalize PAINS-like quinone moieties, a common structural alert.
    "pains_quinone": Chem.MolFromSmarts("O=c1ccccc1=O"),
    # Penalize potentially reactive/unstable iminyl fluorides.
    "iminyl_fluoride": Chem.MolFromSmarts("[#6](=[#7])-[#9]"),
    # Penalize highly reactive acyl halides.
    "acyl_halide": Chem.MolFromSmarts("[#6](=[#8])-[F,Cl,Br,I]"),

}

_SOFT_PENALTY_SUBSTRUCTURES = {
    # Penalize unsubstituted phenols, a metabolic liability.
    "unsubstituted_phenol": Chem.MolFromSmarts("[#6;a]-[#8;H1]"),
    # Penalize potentially unstable N-oxides (as per feedback). Note: a hard
    # charge penalty already exists, but this can be used if that is relaxed.
    "N_oxide": Chem.MolFromSmarts("[#7;+;!$([#7]~[#7])]-[O;-]"),

}

_REWARD_SUBSTRUCTURES = {
    # Known hinge-binding motifs from chemist feedback
    "aminopyrimidine": Chem.MolFromSmarts("c1([#7;H2,H1])ncncc1"),
    "indazole": Chem.MolFromSmarts("c1c2c(cnn2)cccc1"),
    "pyrazolopyrimidine": Chem.MolFromSmarts("c1c2c(cnn2)ncn1"),
    "N_phenylbenzamide": Chem.MolFromSmarts("c1ccccc1NC(=O)c2ccccc2"),
    "imidazopyridazine": Chem.MolFromSmarts("c1n(c2cc[nH]c2)cn1"),
    # Modern medicinal chemistry motifs
    "sulfoximine": Chem.MolFromSmarts("[#16](=[#8])(=[#7])"),
    "alkyl_nitrile": Chem.MolFromSmarts("[#6]-[#6]#[#7]"),
    "trifluoromethyl": Chem.MolFromSmarts("C(F)(F)F"),

}

# Scoring combination
base_score = (
                0.20 * qed
                + 0.15 * logp_score
                + 0.15 * tpsa_score
                + 0.15 * sa_score
                + 0.20 * reward
                + 0.10 * div
                + 0.05 * nov
            )

            # Step 5: Apply soft penalties as a multiplicative factor
            penalty_multiplier = _apply_soft_penalties(mol)
            final_score = base_score
    ...
\end{lstlisting}

\subsubsection*{Objective v4 (excerpt)}
\begin{lstlisting}[language=Python,caption={\texttt{objective\_v4.py}}]
"""Excerpt: objective_v4.py (Human-Agent, kinase, seed 0)."""

_HARD_PENALTY_SUBSTRUCTURES = {
    # Penalize highly reactive acyl halides.
    "acyl_halide": Chem.MolFromSmarts("[#6](=[#8])-[F,Cl,Br,I]"),
    # Penalize metabolically unstable N-F bonds.
    "n_f_bond": Chem.MolFromSmarts("[#7]-[#9]"),

}

_SOFT_PENALTY_SUBSTRUCTURES = {
    # Penalize unsubstituted phenols, a metabolic liability.
    "unsubstituted_phenol": Chem.MolFromSmarts("[#6;a]-[#8;H1]"),
    # Penalize potentially unstable N-oxides.
    "N_oxide": Chem.MolFromSmarts("[#7;+;!$([#7]~[#7])]-[O;-]"),
    # Penalize carboxylates to encourage bioisosteric replacement,
    # but don't filter them out completely as they can be key interactors.
    "carboxylate": Chem.MolFromSmarts("C(=O)[O-]"),

}

_REWARD_SUBSTRUCTURES = {
    # --- Classic/proven kinase hinge-binding motifs ---
    # From new feedback: 2-amino-4-hydroxypyrimidine
    "amino_hydroxy_pyrimidine": Chem.MolFromSmarts("c1(N)nccc(O)n1"),
    "aminopyrimidine": Chem.MolFromSmarts("c1([#7;H2,H1])ncncc1"),
    "indazole": Chem.MolFromSmarts("c1c2c(cnn2)cccc1"),
    "pyrazolopyrimidine": Chem.MolFromSmarts("c1c2c(cnn2)ncn1"),
    "N_phenylbenzamide": Chem.MolFromSmarts("c1ccccc1NC(=O)c2ccccc2"),
    "imidazopyridazine": Chem.MolFromSmarts("c1n(c2cc[nH]c2)cn1"),
    # From new feedback: rigid fused purine-like scaffold
    "fused_imidazo_pyridine_like": Chem.MolFromSmarts("c1ncc2c(n1)cccc2"),

    # --- Modern medicinal chemistry motifs / bioisosteres ---
    "sulfoximine": Chem.MolFromSmarts("[#16](=[#8])(=[#7])"),
    "alkyl_nitrile": Chem.MolFromSmarts("[#6]-[#6]#[#7]"),
    "trifluoromethyl": Chem.MolFromSmarts("C(F)(F)F"),
    # From new feedback: tetrazole as a carboxylate bioisostere
    "tetrazole": Chem.MolFromSmarts("c1nnnn1"),

}

# Scoring combination
base_score = (
                0.25 * qed
                + 0.15 * logp_score
                + 0.10 * tpsa_score
                + 0.25 * sa_score
                + 0.15 * reward
                + 0.05 * div
                + 0.05 * nov
            )

            # Step 5: Apply soft penalties as a multiplicative factor
            penalty_multiplier = _apply_soft_penalties(mol)
            final_score = base_score
    ...
\end{lstlisting}

\subsubsection*{Objective v5 (excerpt)}
\begin{lstlisting}[language=Python,caption={\texttt{objective\_v5.py}}]
"""Excerpt: objective_v5.py (Human-Agent, kinase, seed 0)."""

_HARD_PENALTY_SUBSTRUCTURES = {
    # Penalize highly reactive acyl halides.
    "acyl_halide": Chem.MolFromSmarts("[#6](=[#8])-[F,Cl,Br,I]"),
    # Penalize metabolically unstable N-F bonds.
    "n_f_bond": Chem.MolFromSmarts("[#7]-[#9]"),
    # Penalize Michael acceptors (potential covalent modifiers).
    "michael_acceptor_vinyl_sulfone": Chem.MolFromSmarts("[#6]=[#6]S(=O)=O"),
    "michael_acceptor_acrylamide": Chem.MolFromSmarts("[#6]=[#6]C(=O)N"),
    # Penalize undesirable functionalities.
    "dithiocarboxylate": Chem.MolFromSmarts("[#6](=S)S"),
    "anionic_sulfonate": Chem.MolFromSmarts("S(=O)(=O)[O-]"),

}

_SOFT_PENALTY_SUBSTRUCTURES = {
    # Penalize unsubstituted phenols, a metabolic liability.
    "unsubstituted_phenol": Chem.MolFromSmarts("[#6;a]-[#8;H1]"),
    # Penalize potentially unstable N-oxides.
    "N_oxide": Chem.MolFromSmarts("[#7;+;!$([#7]~[#7])]-[O;-]"),
    # Penalize carboxylates to encourage bioisosteric replacement.
    "carboxylate": Chem.MolFromSmarts("C(=O)[O-]"),

}

_REWARD_SUBSTRUCTURES = {
    # --- Classic/proven kinase hinge-binding motifs ---
    "amino_hydroxy_pyrimidine": Chem.MolFromSmarts("c1(N)nccc(O)n1"),
    "aminopyrimidine": Chem.MolFromSmarts("c1([#7;H2,H1])ncncc1"),
    "indazole": Chem.MolFromSmarts("c1c2c(cnn2)cccc1"),
    "pyrazolopyrimidine": Chem.MolFromSmarts("c1c2c(cnn2)ncn1"),
    "N_phenylbenzamide": Chem.MolFromSmarts("c1ccccc1NC(=O)c2ccccc2"),
    "imidazopyridazine": Chem.MolFromSmarts("c1n(c2cc[nH]c2)cn1"),
    "fused_imidazo_pyridine_like": Chem.MolFromSmarts("c1ncc2c(n1)cccc2"),
    "pyridine_hinge_binder": Chem.MolFromSmarts("c1ccccn1"),
    "pyrimidine_hinge_binder": Chem.MolFromSmarts("c1cncnc1"),

    # --- Modern medicinal chemistry motifs / bioisosteres ---
    "sulfoximine": Chem.MolFromSmarts("[#16](=[#8])(=[#7])"),
    "alkyl_nitrile": Chem.MolFromSmarts("[#6]-[#6]#[#7]"),
    "trifluoromethyl": Chem.MolFromSmarts("C(F)(F)F"),
    "tetrazole": Chem.MolFromSmarts("c1nnnn1"),
    "difluorophenyl": Chem.MolFromSmarts("c1c(F)ccc(F)c1"), # 1,3-difluoro
    "spiro_center": Chem.MolFromSmarts("[*;R2]"), # Atom in exactly two rings

}

# Scoring combination
base_score = (
                0.30 * qed
                + 0.10 * logp_score
                + 0.05 * tpsa_score
                + 0.20 * sa_score
                + 0.10 * fsp3
                + 0.15 * reward
                + 0.05 * div
                + 0.05 * nov
            )

            # Step 5: Apply soft penalties as a multiplicative factor
            penalty_multiplier = _apply_soft_penalties(mol)
    ...
\end{lstlisting}

\subsubsection*{Objective v6 (excerpt)}
\begin{lstlisting}[language=Python,caption={\texttt{objective\_v6.py}}]
"""Excerpt: objective_v6.py (Human-Agent, kinase, seed 0)."""

_HARD_PENALTY_SUBSTRUCTURES = {
    # Penalize highly reactive acyl halides.
    "acyl_halide": Chem.MolFromSmarts("[#6](=[#8])-[F,Cl,Br,I]"),
    # Penalize metabolically unstable N-F bonds.
    "n_f_bond": Chem.MolFromSmarts("[#7]-[#9]"),
    # Penalize Michael acceptors (potential covalent modifiers).
    "michael_acceptor_vinyl_sulfone": Chem.MolFromSmarts("[#6]=[#6]S(=O)=O"),
    "michael_acceptor_acrylamide": Chem.MolFromSmarts("[#6]=[#6]C(=O)N"),
    "michael_acceptor_maleimide": Chem.MolFromSmarts("C1=CC(=O)NC1=O"),
    "michael_acceptor_vinyl_chloride": Chem.MolFromSmarts("[#6]=[#6]-[Cl]"),
    # Penalize undesirable functionalities.
    "dithiocarboxylate": Chem.MolFromSmarts("[#6](=S)S"),
    "anionic_sulfonate": Chem.MolFromSmarts("S(=O)(=O)[O-]"),
    "nitro_group": Chem.MolFromSmarts("[a]-[N+](=O)[O-]"),
    "anionic_nitrogen": Chem.MolFromSmarts("[#7;-]"),

}

_SOFT_PENALTY_SUBSTRUCTURES = {
    # Penalize unsubstituted phenols, a metabolic liability.
    "unsubstituted_phenol": Chem.MolFromSmarts("[#6;a]-[#8;H1]"),
    # Penalize potentially unstable N-oxides.
    "N_oxide": Chem.MolFromSmarts("[#7;+;!$([#7]~[#7])]-[O;-]"),
    # Penalize carboxylates to encourage bioisosteric replacement.
    "carboxylate": Chem.MolFromSmarts("C(=O)[O-]"),

}

_REWARD_SUBSTRUCTURES = {
    # --- Classic/proven kinase hinge-binding motifs ---
    "amino_hydroxy_pyrimidine": Chem.MolFromSmarts("c1(N)nccc(O)n1"),
    "aminopyrimidine": Chem.MolFromSmarts("c1([#7;H2,H1])ncncc1"),
    "indazole": Chem.MolFromSmarts("c1c2c(cnn2)cccc1"),
    "pyrazolopyrimidine": Chem.MolFromSmarts("c1c2c(cnn2)ncn1"),
    "N_phenylbenzamide": Chem.MolFromSmarts("c1ccccc1NC(=O)c2ccccc2"),
    "imidazopyridazine": Chem.MolFromSmarts("c1n(c2cc[nH]c2)cn1"),
    "fused_imidazo_pyridine_like": Chem.MolFromSmarts("c1ncc2c(n1)cccc2"),
    "pyridine_hinge_binder": Chem.MolFromSmarts("c1ccccn1"),
    "pyrimidine_hinge_binder": Chem.MolFromSmarts("c1cncnc1"),
    "phenol_binder": Chem.MolFromSmarts("[c]-[OH1]"),

    # --- Modern medicinal chemistry motifs / bioisosteres ---
    "sulfoximine": Chem.MolFromSmarts("[#16](=[#8])(=[#7])"),
    "alkyl_nitrile": Chem.MolFromSmarts("[#6]-[#6]#[#7]"),
    "trifluoromethyl": Chem.MolFromSmarts("C(F)(F)F"),
    "tetrazole": Chem.MolFromSmarts("c1nnnn1"),
    "difluorophenyl": Chem.MolFromSmarts("c1c(F)ccc(F)c1"), # 1,3-difluoro
    "spiro_center": Chem.MolFromSmarts("[*;R2]"), # Atom in exactly two rings
    "sulfonamide": Chem.MolFromSmarts("S(=O)(=O)N"),
    "sultam": Chem.MolFromSmarts("[#16](=[#8])(=[#8])[#7;R]"), # Cyclic sulfonamide

}

# Scoring combination
base_score = (
                0.30 * qed
                + 0.10 * logp_score
                + 0.05 * tpsa_score
                + 0.20 * sa_score
                + 0.10 * fsp3
                + 0.15 * reward
                + 0.05 * div
                + 0.05 * nov
            )

            # Step 5: Apply soft penalties as a multiplicative factor
            penalty_multiplier = _apply_soft_penalties(mol)
    ...
\end{lstlisting}

\subsection{Distilled Molecular Knowledge (DUD-E)}
\label{sec:appendix_knowledge}

During Surrogate-Expert--Agent optimization, the ExtractAgent consolidates medicinal-chemistry review into a structured knowledge base (\texttt{molecular\_knowledge.json}). Agent-Agent warm-starts from the corresponding Surrogate-Expert--Agent knowledge for the same target and seed. Below we report representative Surrogate-Expert--Agent knowledge for each DUD-E target (prefer seed~0; if empty, the first non-empty seed). Full JSON archives are under \texttt{paper/data/knowledge/benchmark/}.

\subsubsection{Kinase (FAK1)}
Surrogate-Expert--Agent knowledge after feedback rounds (seed~1; 57 entries). Seed~0 was empty or missing, so seed~1 is shown instead.

\paragraph{Structural requirements.}
\begin{itemize}
  \item A gem-difluoro linker can serve as a metabolically stable bioisostere for a carbonyl group.
  \item Chlorinated purine-like moieties, aminopyrimidinones, and uracil-like moieties are classic and effective kinase hinge-binding motifs.
  \item Incorporate privileged kinase scaffolds like imidazo[1,2-b]pyridazine, tetrahydropyrido[2,3-d]pyrimidine, and dihydro-pyrrolo[1,2-a]pyrimidine, which can form multiple strong hydrogen bonds with the hinge.
  \item N-aryl or N-heteroaryl groups are well-suited for occupying the hydrophobic back pocket of the kinase.
  \item Pyridinone and isoquinolinone cores are effective, rigid hinge-binding scaffolds.
  \item Secondary alcohols can serve as effective kinase hinge-binding motifs, acting as both H-bond donors and acceptors.
  \item Side chains appended to the core scaffold can be used to explore interactions in the solvent-exposed or ribose-binding regions.
  \item Single H-bond donors like an aniline N-H can serve as putative, albeit weak, kinase hinge-binders.
  \item Trifluoromethyl (CF3) groups are useful for filling small hydrophobic pockets, improving metabolic stability, and modulating the pKa of adjacent groups.
  \item Use large, rigid hydrophobic cores (e.g., tetrahydroisoquinoline, tetrahydro-naphthalene) to effectively fill the ATP pocket and orient pharmacophores to drive potency.
  \item Utilize rigid, metabolically stable bioisosteres, such as a difluorocyclopropyl group for an isopropyl group, to improve properties while maintaining interactions.
\end{itemize}

\paragraph{Optimization tips.}
\begin{itemize}
  \item Address excessive flexibility by incorporating pharmacophores into more rigid, fused heterocyclic systems (e.g., converting a flexible Ph-CH2-NH-Ph core into an indole or quinoline).
  \item Apply a deconstruction strategy for over-optimized lipophilic hits: replace the problematic group with a simple phenyl to establish a baseline, then systematically rebuild SAR with mono-substituted rings and polar heterocycles.
  \item Assess the chemical stability of novel or unconventional scaffolds early in the optimization process.
  \item Block known metabolic hot-spots (e.g., para-position of a phenyl ring) with inert substituents like fluorine or chlorine.
  \item Convert weak monodentate hinge-binders into more potent bidentate interactions by introducing a nearby H-bond acceptor (e.g., converting an isoquinolinone to a naphthyridinone).
  \item For fragment-like hits, use a growing strategy by adding key pharmacophores (like a hinge-binder) to open vectors on the fragment core.
  \item For hits containing permanent charges, design neutral analogs that preserve the key H-bonding interaction or model the corresponding neutral form to assess affinity retention.
  \item For hits containing reactive groups (e.g., N-haloamides, acyl halides), replace the moiety with a stable analog (e.g., N-H, CF3) and re-evaluate to distinguish true scaffold affinity from reactivity artifacts.
  \item For high-Fsp3 series with low affinity, improve potency by adding larger hydrophobic groups (e.g., phenyl, benzyl) to access the back pocket.
  \item Improve developability of hydrophobic cores by introducing polar atoms, such as replacing a CH with N in an aromatic ring or a CH2 with O in a saturated ring.
  \item Improve metabolic stability and solubility by replacing labile aromatic rings (e.g., phenyl) with less metabolically active heterocycles (e.g., pyridine, pyrazole).
  \item Replace unusual or poorly permeable linkers (e.g., sulfamide) with standard neutral options like amides, reverse amides, sulfonamides, or methylene ethers.
  \item Simplify synthetically complex scaffolds, particularly hydrophobic tails, by replacing them with more accessible fragments (e.g., substituted piperidines, bicyclo[1.1.1]pentanes, bi-aryl systems) to enable faster optimization cycles.
  \item Systematically explore SAR by decorating scaffolds with diverse, small substituents at multiple positions, including peripheral rings (ortho-, meta-, para-) and solvent-facing side chains.
  \item To de-risk high lipophilicity, systematically replace large hydrophobic or poly-halogenated groups with smaller or more polar substituents.
  \item Visualize the docked pose to confirm key interactions (e.g., hinge binding) and to guide the precise placement of new functional groups before synthesis.
  \item When stereocenters are present, plan to synthesize and test all stereoisomers to identify the optimal configuration.
\end{itemize}

\paragraph{Constraints.}
\begin{itemize}
  \item Avoid molecules with non-physiological permanent charges (e.g., phenoxides [O-], sulfonates [SO3-], quaternary amines [N+], protonated heterocycles, deprotonated tetrazoles) to ensure cell permeability and avoid hERG liability.
  \item Avoid structural alerts and reactive functional groups, including nitro groups, N-oxides, N-haloamides (e.g., N-chloroamide), acyl halides (e.g., acyl fluoride), Michael acceptors, and metabolically labile N-O bonds.
  \item Avoid using poly-halogenated aryl groups, which dramatically increase lipophilicity and create developability risks.
  \item Deprioritize computationally intractable or chemically invalid molecules, such as those that are too strained, complex, or contain unsupported charge states.
  \item Deprioritize molecules that are overly large, complex, and lipophilic relative to their potency (poor ligand efficiency), as these are often artifacts of in silico overfitting.
  \item Over-fluorination can be a liability, increasing lipophilicity and creating metabolic red flags.
  \item Penalize or filter out molecules with high flexibility (more than 7-8 rotatable bonds) and strictly filter those with more than 10.
  \item Strictly control lipophilicity, targeting a cLogP below 4 to avoid poor solubility, high plasma protein binding, and P450 inhibition.
  \item Strictly filter out molecules containing sulfonic acids or sulfonates due to permeability issues.
  \item Sulfonamides can pose an off-target liability risk (e.g., carbonic anhydrase).
\end{itemize}

\paragraph{Preferences.}
\begin{itemize}
  \item Actively manage the trade-off between potency and physicochemical properties throughout the design cycle.
  \item Bias generation towards known kinase hinge-binding motifs to increase the hit rate of high-quality binders.
  \item Deconstruct and learn from high-scoring but overly complex or problematic molecules to inspire the design of simpler, more tractable analogs.
  \item Focus generative efforts on the most promising lead series while aggressively de-risking them.
  \item High Fsp3 character is desirable for improving solubility, metabolic stability, and avoiding planarity.
  \item Maintain diverse, lower-risk backup series in parallel with high-risk, high-reward lead series.
  \item Prioritize synthetic accessibility to enable rapid analog synthesis and SAR exploration.
  \item Prioritize the optimization of well-balanced lead candidates with a good profile of potency and drug-like properties.
  \item Reward high ligand efficiency (high potency achieved with low molecular weight and complexity).
  \item Value novel chemotypes for their potential intellectual property (IP) advantages.
\end{itemize}

\paragraph{Insights.}
\begin{itemize}
  \item Chemically-problematic 'activity traps' (e.g., highly reactive, strained, or PAINS molecules) can achieve high docking scores; learn from their binding pose to guide the design of more stable analogs.
  \item Common metabolic liabilities include O-demethylation of methoxy groups, oxidation of saturated rings, glucuronidation/oxidation of benzylic alcohols, para-hydroxylation of N-phenyl rings, and oxidation of benzylic methylenes.
  \item Docking scores can be artificially inflated by strong electrostatic interactions from permanent charges or by the high reactivity of certain functional groups (e.g., acyl halides, N-haloamides) which may act as covalent modifiers.
  \item Failure to generate a docked pose often indicates underlying structural issues like invalid chemistry, severe steric strain, or unsupported charge states.
  \item Generative models can over-optimize for docking scores by creating large, non-drug-like 'Franken-molecules' by stitching together multiple lipophilic fragments; these should be deprioritized and used for inspiration only.
  \item High potency is often achieved at the cost of high lipophilicity, a key trade-off that must be managed, especially in kinase inhibitor design.
  \item High rotatable bond count leads to a high entropic penalty upon binding, poor selectivity, and often high plasma protein binding.
  \item Implement a pre-processing step to check for chemical validity, stability, and successful 3D conformer generation to avoid wasted computation on intractable molecules.
  \item Small fragments with decent binding scores indicate high ligand efficiency and can be excellent starting points for fragment-based design.
\end{itemize}

\subsubsection{GPCR (CXCR4)}
Surrogate-Expert--Agent knowledge after feedback rounds (seed~0; 39 entries).

\paragraph{Structural requirements.}
\begin{itemize}
  \item A protonated basic nitrogen (cationic center) is essential for a salt-bridge interaction with a key acidic residue (e.g., Asp, Glu) in the orthosteric pocket.
  \item Molecules must possess one or two large, rigid, hydrophobic moieties (e.g., polycyclic, aryl, or cycloalkyl) to engage in van der Waals or pi-stacking interactions in adjacent sub-pockets.
  \item A rigid or semi-rigid linker (e.g., methylene, 1,2,4-triazole, piperazine) is critical for correctly orienting pharmacophoric elements and minimizing the entropic penalty of binding.
  \item Preferred scaffolds are rigid or semi-rigid systems, often containing a cationic center, such as: indane, aza-indane, dihydroisoquinoline, benzimidazole, quinazoline, tetrahydro-pyrrolo-pyridine, and tetrahydro-aza-fluorene.
  \item Novel scaffolds like diaminodihydroimidazole and spirocyclic systems are favorable for introducing rigidity and 3D shape complementarity.
  \item Stereochemistry is critical for optimal binding, indicating a well-defined 3D pocket; the active configuration must be determined.
  \item A successful design principle involves connecting two key hydrophobic or heterocyclic recognition motifs with a rigid linker.
\end{itemize}

\paragraph{Optimization tips.}
\begin{itemize}
  \item Optimize for a balanced profile with target properties: Vina score < -9.5 kcal/mol, MW 350-500 Da, cLogP 1-4, QED > 0.6, SA > 0.45, TPSA < 120 \AA{}², and rotatable bonds < 8.
  \item Systematically decorate aromatic and hydrophobic rings on lead scaffolds with small, electronically diverse substituents (e.g., F, Cl, Me, OMe, CF3) to probe pocket boundaries, modulate properties, and block metabolic sites.
  \item Address high lipophilicity by replacing greasy groups (e.g., cyclohexyl, phenyl) with polar bioisosteres (e.g., tetrahydropyran (THP), piperidine, pyridyl) to improve solubility and QED.
  \item Improve metabolic stability by replacing metabolically labile linkers (e.g., thioether, alkyne, N-benzyl) with more stable bioisosteres (e.g., ether, amine, methylene, amide, N-cyclopropylmethyl).
  \item Explore bioisosteric replacements for all key components (hydrophobic groups, basic headgroups, linkers) to fine-tune properties and understand SAR.
  \item When stereocenters are present, synthesize and test all stereoisomers and use modeling to understand the 3D-SAR and predict the active configuration.
  \item Improve solubility by adding small polar substituents to fused aromatic rings or by 'de-flattening' planar systems by introducing sp3-hybridized linkers and cores to increase Fsp3.
  \item Simplify complex or chiral hydrophobic groups with simpler, non-chiral ones (e.g., adamantyl, cyclohexyl) to improve synthetic tractability without sacrificing affinity.
  \item Introduce potential hydrogen-bonding groups (e.g., a hydroxyl) on linkers to probe for new interactions with nearby residues.
  \item Deconstruct complex, high-scoring molecules ('molecular blobs') to identify the minimal pharmacophore required for affinity, then rebuild with better properties.
\end{itemize}

\paragraph{Constraints.}
\begin{itemize}
  \item Avoid molecules outside the optimal molecular weight range of 350-500 Da; enforce hard filters against MW > 550 Da and a soft filter against MW < 250 Da.
  \item Strictly limit molecules to one or two basic centers; avoid structures with more than two predicted positive charges at physiological pH due to poor permeability and off-target risks like phospholipidosis.
  \item Deprioritize or filter out molecules with poor predicted drug-like properties, such as QED < 0.6, synthetic accessibility (SA) < 0.4, TPSA > 140 \AA{}², or cLogP > 5.
  \item Avoid chemically unstable or metabolically labile functionalities like thioethers (oxidation risk), anilines (toxicity/metabolism risk), alkynes, and certain exocyclic imines (hydrolysis risk).
  \item Avoid 'molecular obesity': molecules with excessive size, lipophilicity, complexity, or flexibility (e.g., more than 8-10 rotatable bonds, or more than two large hydrophobic regions).
  \item Avoid overly planar, fused aromatic systems ('brick dust' candidates) which have liabilities related to poor solubility, aggregation, hERG binding, and potential DNA intercalation.
  \item Discard scaffolds that consistently fail to dock, suggest severe steric clashes, or possess problematic chemistry like geometrically strained systems.
  \item Permanent charges (e.g., quaternary amines) are a major permeability liability and should be avoided.
  \item Forbid synthetically intractable structures, such as those with SA scores below a hard threshold of 0.4.
  \item Explicitly forbid or heavily penalize long, flexible polyamine chains due to high entropic penalty, non-specific binding, and poor ADME properties.
\end{itemize}

\paragraph{Preferences.}
\begin{itemize}
  \item Employ a multi-parameter optimization approach, balancing potency with ADME properties (QED, SA, MW, cLogP, TPSA), rather than focusing solely on affinity.
  \item Shift the primary optimization goal from raw docking score (potency) to Ligand Efficiency (LE = Vina / Heavy Atom Count) to reward smaller, more efficient binders and avoid molecular obesity.
  \item Prioritize potent, rigid scaffolds with good 3D shape complementarity (high Fsp3) to reduce the entropic penalty of binding and avoid the liabilities of flat molecules.
  \item Favor molecules with manageable synthetic accessibility (SA > 0.45), especially in early optimization stages.
  \item Focus optimization efforts on decorating and exploring analogs of validated, high-performing lead scaffolds ('lead optimization' strategy).
  \item Maintain a backup strategy by exploring alternative, structurally distinct scaffolds in parallel with lead optimization.
\end{itemize}

\paragraph{Insights.}
\begin{itemize}
  \item A successful pharmacophore model for this GPCR target consists of a cationic anchor (forming a salt-bridge) connected by a rigid linker to one or two hydrophobic/heterocyclic regions that engage adjacent sub-pockets.
  \item Common failure modes include 'potency traps' or 'greaseballs' (high affinity but poor properties due to excessive size/lipophilicity), 'molecular blobs' (synthetically intractable complexity), and 'brick dust' (overly planar systems with poor solubility).
  \item Flexible linkers or side chains incur a significant entropic penalty upon binding, reducing affinity compared to more rigid analogues.
  \item Small fragments lacking sufficient hydrophobic bulk are unable to achieve high affinity, confirming that both the basic anchor and a large hydrophobic component are required.
  \item Learnings and Structure-Activity Relationships (SAR) from a lead series can often be successfully transferred to a structurally related backup scaffold.
  \item A strong inverse correlation often exists between raw docking score and developability, requiring a focus on balanced properties rather than just affinity.
\end{itemize}

\subsubsection{Nuclear receptor (MCR)}
Surrogate-Expert--Agent knowledge after feedback rounds (seed~0; 48 entries).

\paragraph{Structural requirements.}
\begin{itemize}
  \item A central phenol flanked by two phenyl rings (a terphenyl-like system) is a classic nuclear receptor pharmacophore, with the hydroxyl serving as a key H-bond anchor.
  \item Favor established nuclear receptor scaffolds (e.g., tetralones, quinolinones) or novel rigid, fused heterocyclic systems such as those containing pyridinone, imidazole, benzofuran/benzofuranone, isoxazole, lactam, benzimidazole, dibenzothiophene S-oxide/dioxide, or fused sulfonamides.
  \item Incorporate potent hydrogen bond donors and acceptors (e.g., phenols, ketones, amides, sulfones, sulfoxides, nitriles) and specific heterocycles, as they are critical for binding to nuclear receptors.
  \item Prioritize rigid, multi-cyclic, spirocyclic, or sp3-rich scaffolds to minimize the entropic penalty of binding.
  \item The Aryl-Alkyl-Sulfone-Alkyl-OH motif is a highly successful linear scaffold combining hydrophobic and H-bonding features.
  \item Utilize scaffolds with hydrophobic bulk (e.g., fluorophenyl, tert-butyl, cyclohexyl groups, multiple aromatic rings) to effectively fill the large, lipophilic binding pocket of nuclear receptors.
\end{itemize}

\paragraph{Optimization tips.}
\begin{itemize}
  \item Abstract the 3D pharmacophore from synthetically complex but high-affinity hits and replicate it on a simpler, more accessible rigid core.
  \item Benchmark activity and properties of acidic heterocycles (e.g., tetrazole) by synthesizing analogs with other acidic groups (e.g., triazole) or a simple carboxylic acid.
  \item Block metabolism of key groups, such as by adding fluorine or methyl substituents ortho to a phenolic hydroxyl to prevent oxidation.
  \item Confirm the role of key H-bonding groups like hydroxyls by synthesizing masked analogs (e.g., methyl ethers) to test for activity loss.
  \item Explore modifications to key functional groups to probe for new interactions, such as converting a ketone to a hydroxyl, oxime, or difluoromethylene (CF2).
  \item Grow molecules from simple, compact cores (e.g., quinolinone) by adding substituents to probe deeper into the ligand-binding domain.
  \item Improve ADME properties by managing lipophilicity; strategies include replacing phenyl rings with less-lipophilic heteroaromatics (e.g., pyridine, pyrimidine), adding small polar groups (e.g., OH, OMe), or saturating double bonds to increase sp3 character.
  \item Improve metabolic stability by replacing labile groups or linkers (e.g., acetal, thioether, dihydropyridine) with stable bioisosteres (e.g., ether, methylene, alcohol, stable aromatic rings).
  \item Introduce alkyl groups at the ortho-positions of biaryl systems to force a specific dihedral angle and conformation.
  \item Introduce small alkyl groups (e.g., methyl, ethyl, cyclopropyl) on saturated rings to probe for additional hydrophobic contacts.
  \item Neutralize charged functional groups to improve permeability by capping them (e.g., as methyl esters) or replacing them with neutral bioisosteres (e.g., carboxylate to tetrazole/hydroxyl; anionic sulfonamide to N-H/N-alkyl analog; pyridinium to pyridine).
  \item Replace reactive functional groups like acyl fluorides with stable bioisosteres that mimic their electronic properties, such as a nitrile, CF3, or amide.
  \item Rigidify flexible multi-ring systems by connecting rings with stable bridges (e.g., oxygen, methylene) to form fused cores like dibenzofurans or fluorenes, thereby locking in an active conformation.
  \item Systematically probe aromatic rings with small electron-donating (Me, OMe) and withdrawing (halogens, CN) groups at various positions (ortho, meta, para) to map pockets, block metabolism, and find new interactions.
  \item Transfer successful substitution patterns from simple molecules onto more complex, 3D scaffolds ('scaffold hopping').
  \item Utilize bioisosteric replacement extensively to modulate properties: ester to amide/1,2,4-oxadiazole/nitrile; carboxylic acid/carboxylate to tetrazole/acylsulfonamide/hydroxyl; sulfone to sulfonamide/sulfoximine; sulfoxide to sulfone/ketone/amide; lactone to lactam/cyclic ether/carbocycle; imidazole to triazole/pyrazole/oxazole; benzofuranone to isoindolinone; thiophene to furan/thiazole/phenyl/pyridine; Michael acceptor to tert-butyl/cyclohexyl/phenyl; pyridinium to pyridine.
\end{itemize}

\paragraph{Constraints.}
\begin{itemize}
  \item Avoid known structural alerts and reactive functionalities, including Michael acceptors, N-oxides, N-F bonds, imines, acyl fluorides, N-chloroamines, terminal alkenes, epoxides, aziridines, and diazene derivatives.
  \item Avoid large, flat, extended conjugated, or overly planar systems (e.g., >4 fused aromatic rings) due to potential for poor solubility, non-specific binding (PAINS), hERG liability, and DNA intercalation.
  \item Avoid molecules with high nitrogen content, which can lead to poor permeability and off-target effects.
  \item Avoid or immediately replace metabolically liable groups, such as acetal/thioacetal linkers, esters, lactones, thioethers, and dihydropyridines.
  \item Avoid or neutralize charged groups (e.g., carboxylates, phenolates, anionic sulfonamides, pyridiniums) and zwitterions to ensure good cell permeability and oral bioavailability; penalize non-physiological formal charges.
  \item Deprioritize molecules that fail to dock, as they may be too large, have a poor conformational fit, or contain invalid chemistry.
  \item Enforce property filters to penalize molecules with high molecular weight (>500-550 Da), high lipophilicity (cLogP > 5), high rotatable bond count (>10), poor synthetic accessibility (SA score < 0.4), or excessive complexity (>8 rings or >4 stereocenters).
  \item For lead optimization, avoid generating overly simple fragments by setting minimum complexity thresholds (e.g., MW > 200 Da or heavy atom count > 15).
\end{itemize}

\paragraph{Preferences.}
\begin{itemize}
  \item Balance a strong predicted potency (e.g., Vina < -7.0 kcal/mol) with overall developability, prioritizing candidates with favorable physicochemical properties (e.g., QED > 0.5, SA > 0.4).
  \item Favor non-planar, 3D, sp3-rich scaffolds over flat, flexible, or overly simple molecules to improve selectivity and physicochemical properties.
  \item For cell permeability, prioritize molecules with neutral phenols over their corresponding anionic phenolates, even if the anion shows a better docking score.
  \item Prioritize synthetically accessible and tunable scaffolds over those with high synthetic complexity, even if the latter have high predicted affinity.
  \item Systematically explore Structure-Activity Relationships (SAR) around the most promising scaffolds.
\end{itemize}

\paragraph{Insights.}
\begin{itemize}
  \item A well-substituted aromatic ring can be an effective fragment-like starting point for optimization.
  \item Alkyl sulfones and primary alcohols are effective and drug-like hydrogen bond acceptors and donors, respectively.
  \item Basic nitrogen atoms within heterocycles (e.g., imidazole, pyridine) can be used to improve aqueous solubility and provide a handle for salt formulation.
  \item Fluorine substitution on aromatic rings can enhance binding affinity, block metabolism, and improve physicochemical properties.
  \item High lipophilicity (cLogP), while often good for binding in greasy pockets, can lead to poor aqueous solubility and high plasma protein binding, negatively impacting ADME and bioavailability.
  \item Nuclear receptor ligand-binding pockets are typically large and lipophilic, favoring rigid molecules with significant hydrophobic bulk.
  \item Phenolic hydroxyls are critical pharmacophoric features in many nuclear receptor ligands but are also significant metabolic liabilities subject to oxidation and glucuronidation.
  \item Rigidity enhances binding affinity by pre-organizing the molecule for the target and minimizing the entropic penalty associated with conformational restriction upon binding.
  \item Simple, compact, and synthetically tractable drug-like scaffolds (e.g., quinolinone) can serve as excellent starting points for elaboration into more potent molecules.
  \item Sulfoxide groups are strong H-bond acceptors but can be metabolic liabilities and introduce chirality.
  \item Tetrazole is an effective neutral bioisostere for a carboxylic acid.
  \item The nitrile group is a potent, metabolically stable, and linear hydrogen bond acceptor.
  \item The trifluoromethyl (CF3) group increases lipophilicity and can act as a weak hydrogen bond acceptor, often making favorable contacts in hydrophobic pockets.
\end{itemize}

\subsubsection{JSON Structure of the Knowledge Base}
\begin{itemize}
    \item \textbf{Structural Requirements:} Scaffold preferences and functional group requirements
    \item \textbf{Optimization Tips:} Target ranges and strategies for property improvement
    \item \textbf{Constraints:} Hard boundaries such as molecular weight limits ($300 < \text{MW} < 600$)
    \item \textbf{Preferences:} Subjective design philosophies of the medicinal chemist
    \item \textbf{Insights:} General domain-specific best practices
\end{itemize}

\subsubsection{Design Principles of the CodeAgent}
\begin{itemize}
\item \textit{Score Normalization:} Every generated objective function is constrained to return normalized values in the range $[0, 1]$. We define $1$ as the optimal representation of the chemist's target profile, while $0$ indicates a complete failure to meet essential criteria.

\item \textit{Operational Robustness:} To handle the ``messiness'' of the generated chemical knowledge, the agent implements granular, per-molecule error handling. Any invalid SMILES strings or molecules that violate hard constraints (e.g., radical species or forbidden substructures) are automatically assigned a score of $s = 0$ to prevent runtime failures in the training loop.

\item \textit{Automated Self-Verification:} After code generation, the agent executes a self-correction protocol. This involves a static analysis of the code to detect logical flaws and a verification step to ensure that all constraints stored in the persistent knowledge state in $\mathcal{K}$ are preserved in the updated function.
\end{itemize}

\subsubsection{Hallucination Mitigation and Robustness}

To address concerns regarding LLM hallucinations in a low-context chemical space, the system employs three layers of defense:

\begin{itemize} \item \textbf{Domain-Specific Persona Conditioning:} Every agent in the pipeline is initialized with an "Expert Medicinal Chemist" persona. This conditioning restricts the LLM's latent space to professional terminology and evidence-based reasoning, significantly reducing the likelihood of generic or chemically nonsensical outputs.

\item \textbf{Structured Reasoning via Chain-of-Thought (CoT):} We enforce a rigorous Chain-of-Thought extraction process. Agents are required to deconstruct the chemist's feedback, identify underlying constraints, and justify their logic before outputting any structured JSON or code. This transparency allows the system to detect logical inconsistencies before they propagate to the generative model.

\item \textbf{Deterministic Code-Level Grounding:} The most critical safeguard is the decoupling of generative logic from molecular execution. The LLM does not generate SMILES strings directly; instead, it generates the \textit{objective scoring logic}. The actual chemical evaluation is handled by deterministic, industry-standard tools (e.g., RDKit) called within the Python-based objective function. This ensures that rewards are grounded in physical reality and verified chemical informatics rather than statistical prediction.
\end{itemize}

\subsection{Ablation Study on Agentic System}

\begin{figure}[t]
\begin{center}
\includegraphics[scale=0.17]{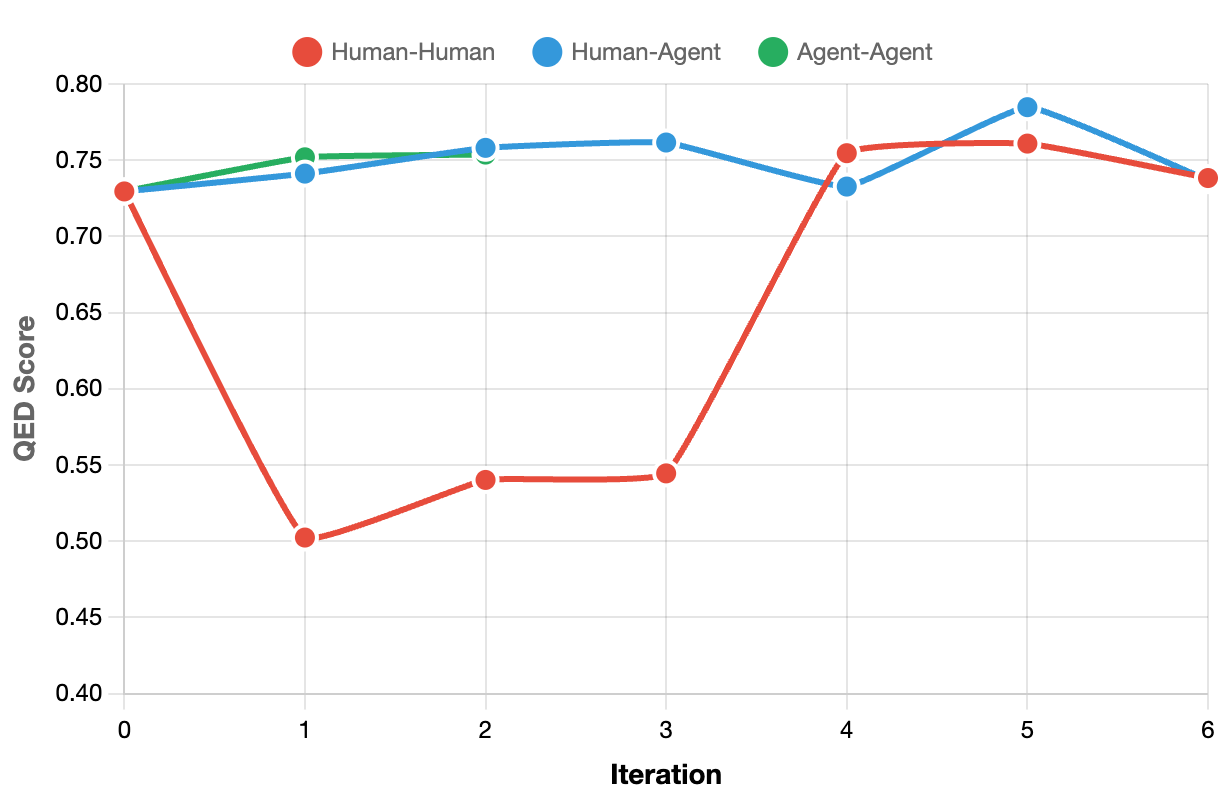}
\end{center}
\caption{\textbf{QED (Quantitative Estimate of Drug-likeness):} A comprehensive drug-likeness score ranging from 0 to 1 that combines eight molecular descriptors including molecular weight, logP, hydrogen bond donors/acceptors, and structural complexity. Higher scores indicate greater pharmaceutical attractiveness.
}
\label{fig:qed_agentic_system}
\end{figure}

\begin{figure}[t]
\begin{center}
\includegraphics[scale=0.17]{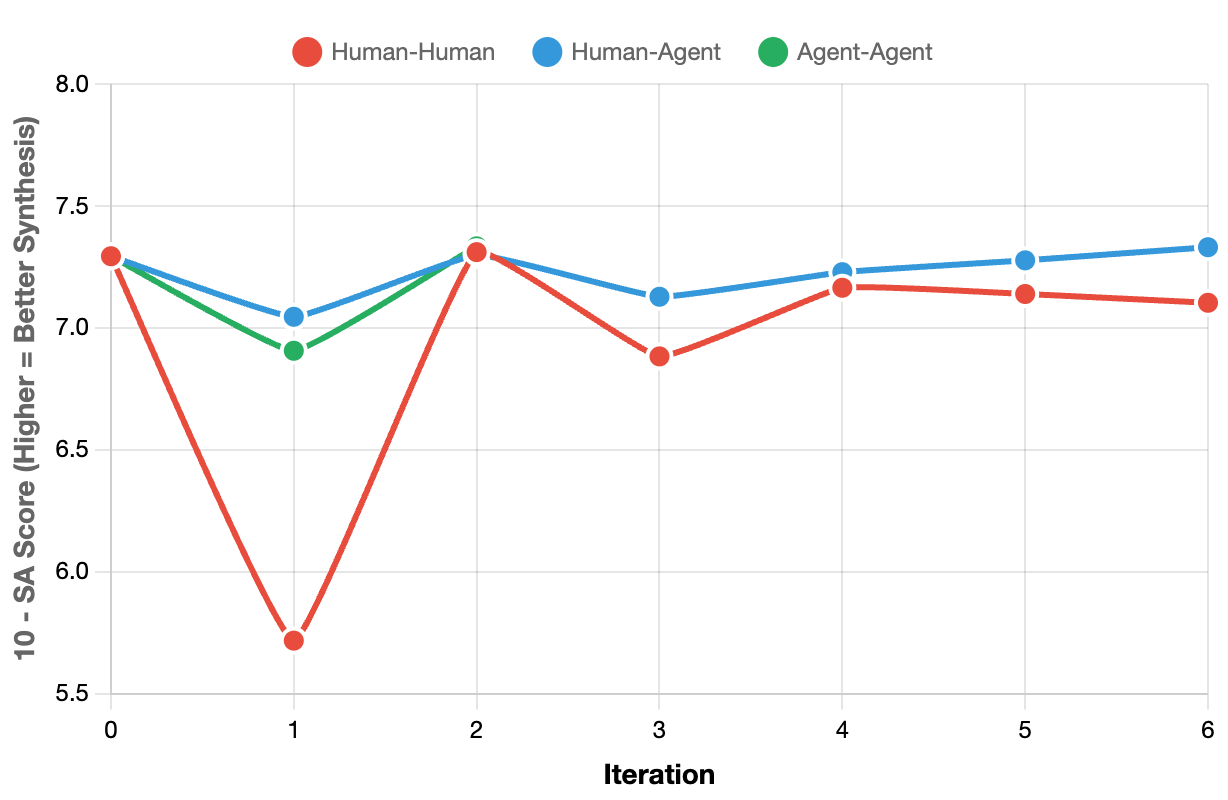}
\end{center}
\caption{\textbf{SA (Synthetic Accessibility):} A computational metric predicting synthetic difficulty on a scale from 1 (very easy) to 10 (very difficult). Lower scores indicate molecules that are easier to synthesize in practice, which is crucial for drug development feasibility.}
\label{fig:sa_agentic_system}
\end{figure}

Figure~\ref{fig:qed_agentic_system} and Figure~\ref{fig:sa_agentic_system} show QED and SA scores across deployment iterations on the proprietary cancer-lead campaign. All models began from the same baseline, but the Human-Agent and Agent-Agent configurations showed more consistent trajectories than the Human-Human approach. Since initial QED and SA scores were already high, feedback from medicinal chemists focused on target-specific information. Our optimization therefore prioritized improving docking scores while maintaining other metrics. The process stops when chemists approve generated candidates for in vitro testing.










\end{document}